\documentclass[sigconf,screen]{acmart}
\AtBeginDocument{%
  }

\usepackage{algorithm}
\usepackage{algorithmic}
\usepackage{enumitem}
\usepackage{multirow}
\usepackage{fontawesome5}
\usepackage{diagbox}
\usepackage{colortbl}
\usepackage{makecell}
\usepackage{mdframed}
\usepackage{xspace}
\usepackage{subcaption}
\usepackage{threeparttable}
\usepackage{url}
\usepackage{booktabs}
\usepackage{graphicx}
\usepackage{xcolor}
\usepackage{pifont}
\usepackage{balance}
\usepackage{amsmath}
\usepackage{booktabs} 
\usepackage{array} 
\usepackage{CJKutf8}
\usepackage{adjustbox} 
\usepackage[most]{tcolorbox} 
\usepackage{tabularx}
\usepackage{array}
\usepackage{graphicx}
\usepackage{dashrule}
\usepackage{etoolbox}
\usepackage{CJKutf8}
\usepackage{graphicx}
\usepackage{xcolor}
\usepackage{tikz}
\usetikzlibrary{calc,positioning,fit,shadows.blur}
\usepackage[most]{tcolorbox}
\usepackage{varwidth}

\copyrightyear{2026}
\acmYear{2026}
\setcopyright{cc}
\setcctype{by}
\acmConference[KDD '26]{Proceedings of the 32nd ACM SIGKDD Conference on Knowledge Discovery and Data Mining V.2}{August 09--13, 2026}{Jeju Island, Republic of Korea}
\acmBooktitle{Proceedings of the 32nd ACM SIGKDD Conference on Knowledge Discovery and Data Mining V.2 (KDD '26), August 09--13, 2026, Jeju Island, Republic of Korea}
\acmDOI{10.1145/3770855.3818068}
\acmISBN{979-8-4007-2259-2/2026/08}

\newcommand{\model}{\texttt{UrbanAgent}}

\definecolor{myPurple}{HTML}{92358e}
\definecolor{myblue}{HTML}{489cd0}

\begin{document}

\newcommand{\PlainTitle}{Multi-Agent Collaborative Reasoning with Tool-Augmented Evidence for Urban Region Profiling}

\title[UrbanAgent]{
Multi-Agent Collaborative Reasoning with \\ Tool-Augmented Evidence for Urban Region Profiling}

\author{Xixuan Hao}
\affiliation{%
  \institution{The Hong Kong University of Science and Technology (Guangzhou)}
  \city{Guangzhou}
  \country{China}}
\email{xhao390@connect.hkust-gz.edu.cn}

\author{Yutian Jiang}
\affiliation{%
  \institution{The Hong Kong University of Science and Technology (Guangzhou)}
  \city{Guangzhou}
  \country{China}}
\email{yjiang194@connect.hkust-gz.edu.cn}

\author{Jiabo Liu}
\affiliation{%
  \institution{The Hong Kong University of Science and Technology (Guangzhou)}
  \city{Guangzhou}
  \country{China}}
\email{jliu933@connect.hkust-gz.edu.cn}

\author{Yihang Yang}
\affiliation{%
  \institution{University of Washington}
  \city{Seattle}
  \country{USA}}
\email{yheason6@uw.edu}

\author{Guangyin Jin}
\affiliation{%
  \institution{Chang'an University}
  \city{Xi'an}
  \country{China}}
\email{jinguangyin96@foxmail.com}

\author{Song Gao}
\affiliation{%
  \institution{University of Wisconsin - Madison}
  \city{Madison}
  \country{USA}}
\email{song.gao@wisc.edu}

\author{Yuxuan Liang}
\authornote{Corresponding author. Email: yuxliang@outlook.com}
\affiliation{%
  \institution{The Hong Kong University of Science and Technology (Guangzhou)}
  \city{Guangzhou}
  \country{China}
}
\email{yuxliang@outlook.com}

\renewcommand{\shortauthors}{Xixuan Hao et al.}
\begin{abstract}
Urban region profiling constitutes a core problem in urban computing, supporting applications such as population estimation, economic assessment, and environmental monitoring. Existing methods typically formulate this task as multimodal representation learning, fusing heterogeneous urban data—e.g., satellite imagery, points of interest, textual descriptions, and 3D building information—into latent embeddings for prediction. However, these approaches are largely correlation-driven, assume cross-modal consistency, and rely on static pipelines, which limit their robustness in heterogeneous or unseen urban regions.
We propose \model, an agentic framework that reframes urban region profiling as a reasoning-driven inference problem. \model~instantiates an independent agent for each data modality and performs structured multi-agent collaborative reasoning to explicitly address cross-modal inconsistencies rather than absorbing them into a single representation. 
In addition, \model~extends indicator prediction as a closed-loop process of active evidence acquisition and iterative reasoning, enabling agents to verify uncertain inferences through tool-augmented retrieval of external knowledge optimized via reinforcement learning. Extensive experiments on global urban datasets for Carbon emissions, GDP, and Population estimation show that \model~consistently outperforms existing baselines, achieving an average improvement of 8.1\% in $R^2$
, and exhibiting strong generalization performance in unseen-city settings.
\end{abstract}

\maketitle

\section{Introduction}
\label{sec:introduction}
Understanding urban regions from data poses a fundamental challenge in urban computing~\cite{zou2025deep}, with broad applications in population estimation, economic assessment, and environmental monitoring. With the growing availability of heterogeneous urban data, recent studies have explored \textit{urban region profiling}~\cite{hao2025unlockingsurvey}, aiming to infer region-level attributes by jointly leveraging multimodal data including satellite imagery, points of interest (POIs), textual descriptions and 3D building information. In recent years, most existing approaches formulate this task as a representation learning problem~\cite{hao2025unlockingsurvey}, where multimodal inputs are encoded into latent representations and mapped to target attributes through Deep Learning (DL) techniques~\cite{geohg,yeh2020using,urbanclip,urbanvlp,regiondcl,musecl,recp,romer,satcle}.

\begin{figure}[!t]
  \centering
  \includegraphics[width=\linewidth]
  {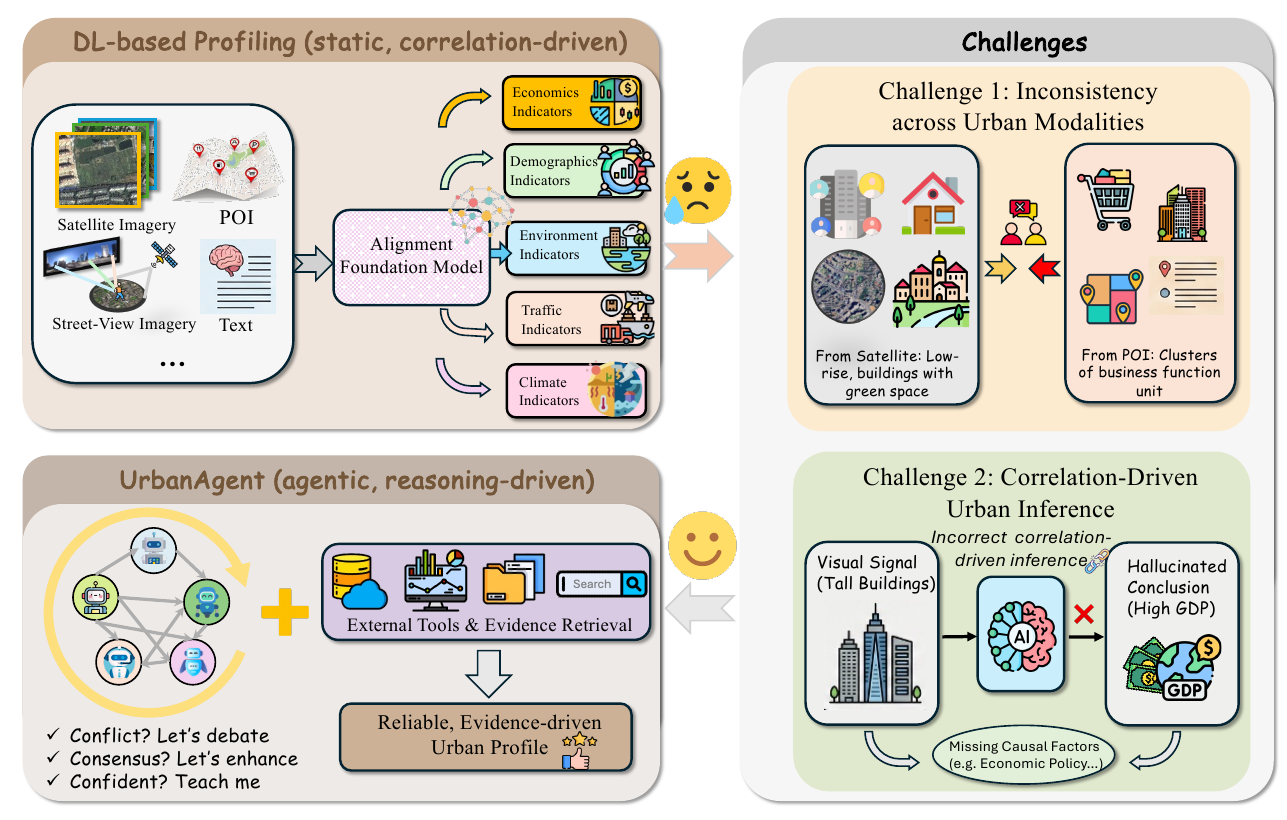}
  \vspace{-2.5em}
  \caption{The concepts and challenges of urban region profiling. Our \model~is a pioneering research on using multi-agent collaborative reasoning for this task.}
  \label{fig:intro1}
  \vspace{-1em}
\end{figure}

While effective in specific settings, this paradigm exhibits inherent limitations when applied to complex urban systems~\cite{li2025urbancomputingsurvy}. In particular, prior DL–based profiling methods rely on latent fusion spaces, assume strong modality schema, and operate within static alignment pipelines. As a result, these models are primarily \textbf{correlation-driven}: \textit{they learn statistical associations from the training data, but lack the ability to reason about conflicting evidence, adapt to new contexts, or verify uncertain inferences}. These limitations become especially pronounced in heterogeneous or previously unseen urban regions~\cite{zhong2024urbancross,geohg}.

In this work, we argue that the challenges of urban region profiling are not merely technical issues of representation learning, but reflect a deeper mismatch between the learning paradigm and the complex nature of urban data. 
As shown in Figure~\ref{fig:intro1}, we identify two fundamental challenges that motivate a shift from static, correlation-driven modeling toward agentic, reasoning-driven urban inference:

    \ding{70}  \textbf{Inconsistency across Urban Modalities}. Urban regions are observed through multiple heterogeneous data modalities, such as satellite imagery, POI records, 3D buildings and texts. These modalities capture different facets of urban reality, including physical form, functional usage, and semantic perception, thereby often providing inconsistent signals for the same region. For instance, satellite imagery may depict a low-density residential area, while POI data indicate dense commercial activities; 3D building models may show only a few high-rise structures, and textual descriptions may label the same area as an “emerging business district.” Such discrepancies are common in real-world cities and reflect their mixed-use nature rather than noise. However, most existing multimodal profiling methods assume that heterogeneous signals can be aligned and fused into a unified representation. When modalities are inconsistent, misleading signals are combined without being explicitly identified, causing cross-modal inconsistencies to be absorbed into the fused representation, ultimately leading to inaccurate and unreliable urban assessments.

\ding{70}  \textbf{Correlation-Driven Urban Inference}. Urban region profiling models largely rely on correlations learned from data to infer complex attributes. In practice, easily observable cues, including building height, night-light intensity, POI density, are often associated with high-level attributes like gross domestic product (GDP), population, or urban function. While such correlations may hold on average, they are not universally reliable across diverse urban contexts. Tourism districts, for example, may exhibit strong night-time activity without high resident population, whereas building height and skyline density may serve as unreliable proxies for GDP across heterogeneous urban contexts. 
Nevertheless, correlation-driven models tend to apply learned associations uniformly across regions, even when local contexts differ from those seen during training. This can lead to confident yet weakly supported predictions, particularly in heterogeneous or previously unseen regions.

\begin{table}[!t]  
\centering
\small
\caption{Paradigm-level comparison between conventional DL-based urban region profiling methods and \model.}
\begin{tabular}{lcc}
\toprule
\textbf{Aspect} 
& \textbf{DL-based Profiling} 
& \textbf{UrbanAgent (Ours)} \\
\midrule
Fusion Space 
& Latent Embedding 
& \textbf{Semantic Reasoning} \\

Schema Dependency 
& Strong 
& \textbf{Weak / None} \\

Processing Pipeline 
& Static 
& \textbf{Agentic} \\

Adaptability 
& Limited 
& \textbf{General-purpose} \\

Modeling Paradigm 
& Correlation-driven  
& \textbf{Reasoning-driven} \\
\bottomrule
\end{tabular}
\label{tab:intro_summarize}
\end{table}

Overall, these challenges suggest that urban region profiling cannot be adequately framed as a static feature fusion task. Instead, it necessitates a \textit{reasoning-centric} paradigm that can operate over heterogeneous urban evidence, relaxing strong schema assumptions, and adapting inference behavior to diverse urban contexts, rather than relying solely on correlations learned from training data. Table~\ref{tab:intro_summarize} distills these challenges and motivates the need for a reasoning-centric paradigm.

In this paper, we propose \textbf{\model}, an agentic framework for multimodal urban region profiling. \model~addresses modality inconsistency by explicit multi-agent collaborative reasoning with structured inter-agent relations, rather than jointly embedding all inputs into a unified latent representation. This design allows inconsistencies across urban modalities to be addressed explicitly instead of implicitly absorbed. 
In addition, \model~mitigates correlation-driven inference by extending the urban indicator prediction as a closed-loop process of active evidence acquisition and iterative reasoning. 
Rather than uniformly applying correlations learned from training data, \model~enables inference behavior to adaptively incorporate external, grounded knowledge as complementary evidence, thereby reducing reliance on brittle surface-level cues in heterogeneous or previously unseen urban regions. Our contributions are summarized as follows:

\begin{itemize}
    [leftmargin=*]
    \item  \textit{Multi-Agent Collaborative Reasoning Framework}. We propose \model, an agentic AI framework that reframes multimodal urban profiling as a semantic reasoning problem rather than a latent embedding task. By modeling each
modality as an independent agent, \model~enables flexible cross-modal interaction without requiring strong schema
assumptions.
    \vspace{0.2em}
    \item \textit{Tool-Augmented Agentic Learning}. \model~introduces a tool-augmented agentic learning paradigm that empowers agents to actively acquire and verify external evidence through tool use. By extending urban indicator prediction process into a closed-loop reasoning process and optimizing tool-use via reinforcement learning, our approach effectively mitigates correlation-driven inference and over-reliance on LLM internal priors.
    \item  \textit{Extensive Empirical Studies}. We conduct comprehensive experiments on global urban datasets covering Carbon emissions, GDP, and Population, under both in-domain and unseen-city settings. 
    The results show that \model~consistently outperforms existing DL and multi-agent baselines by 8.1\% on the $R^2$ metric, validating the effectiveness of collaborative reasoning and evidence-driven inference for urban region profiling.
\end{itemize}

\begin{figure*}[ht]
  \centering
  \includegraphics[width=\linewidth]
  {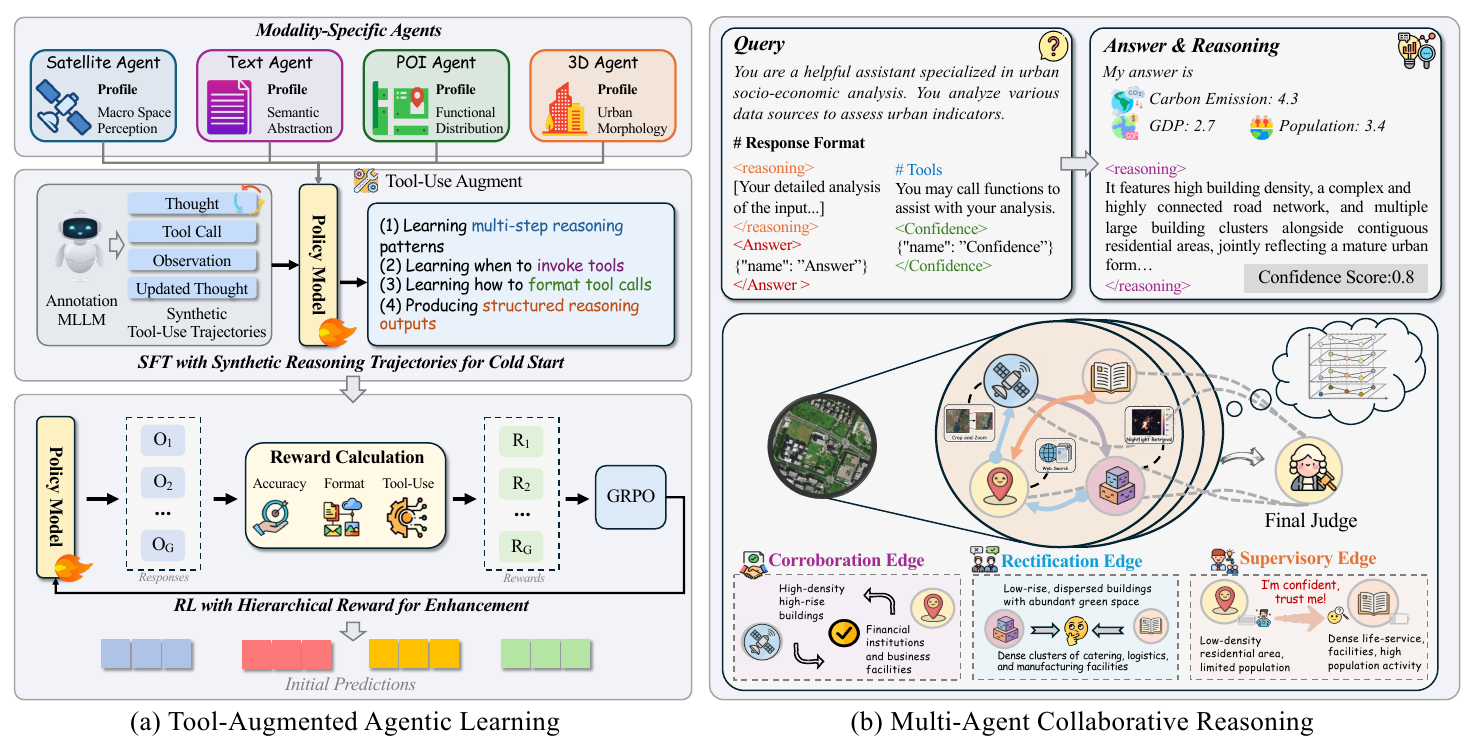}
  \vspace{-2.2em}
  \caption{The overall Framework of \model.}
  \label{fig:framework}
  \vspace{-1em}
\end{figure*}

\vspace{-0.5em}
\section{Preliminary}
\subsection{Problem Formulation}

Formally, for a dataset of $N$ urban regions $\{u_i\}_{i=1}^N$,
we are given a set of multimodal inputs
$\mathcal{X}_{u_i} = \{x_{u_i}^{\mathrm{sat}},\, x_{u_i}^{\mathrm{text}},\, x_{u_i}^{\mathrm{poi}},\, x_{u_i}^{\mathrm{3d}}\}$,
where $x_{u_i}^{\mathrm{sat}} \in \mathbb{R}^{H \times W \times C}$ denotes a satellite image patch with spatial resolution $H \times W$ and $C$ channels;
$x_{u_i}^{\mathrm{text}}$ represents textual information describing the region;
$x_{u_i}^{\mathrm{poi}} \in \mathbb{R}^{P_{u_i} \times M}$ represents POI category information for region $u_i$, where $P_{u_i}$ denotes the number of POI instances in the region, and each row is an $M$-dimensional one-hot vector over the POI category set;
$x_{u_i}^{\mathrm{3d}} = \{(\mathcal{P}_i, h_i)\}$, where $\mathcal{P}_i$
denotes the planar footprint polygon of region $i$, and
$h_i$ denotes its measured height.
The Urban Region Profiling (URP) problem is therefore to learn a mapping 
$\mathcal{F}: \mathcal{X}_{u_i} \rightarrow \text{Y}_{u_i},$
to infer $D$ urban indicators $\text{Y}_{u_i} \in \mathbb{R}^{D}$.

\vspace{-0.5em}
\subsection{Related Work}

\subsubsection{Urban Region Profiling}
URP aims to predict region-level socioeconomic indicators from heterogeneous urban observations, with the central challenge of capturing the multi-faceted nature of urban environments. Early studies learned region representations from human mobility and POIs~\cite{region2vec,mte,wen2024demo2vec,recp,romer,hrep,mvgrnet,geohg}, characterizing human activity patterns in urban space. 
Some work incorporated physical environment cues via satellite or street-view imagery to model \textit{what the city looks like}~\cite{liu2025citylens,regionencoder,jean2019tile2vec,knowcl,pgsimclr,geohg,li2023point}.
Follow-up studies further integrate visual, human-activity, and semantic signals to inject higher-level urban concepts~\cite{muhlematter2025urbanfusion,zhao2025graphjcl,urbanclip,urbanvlp}. 
More recently, self-supervised pretraining and multimodal foundation models have been explored for urban analysis~\cite{xiao2024refound,yong2024musecl,liu2026urbanmoe,urbanln,cao2026urbanmmcl}.
However, naively compressing potentially conflicting modalities into a unified representation can cause semantic dilution and logical gaps.
In this work, we propose a structured reasoning framework that explicitly preserves and reconciles heterogeneous urban modalities, enabling verifiable and interpretable urban region profiling.

\vspace{-0.2em}
\subsubsection{Agentic Reinforcement Learning (RL)}

Agentic RL reframes LLMs as operational agents capable of multi-step reasoning and environment interaction~\cite{agenticrl-survey}, supported by stable policy optimization methods such as GRPO~\cite{guo2025deepseekr1,trafficr1}. By integrating a loop of reasoning–action–observation~\cite{react,zhang2025agentrl}, agents can invoke external tools to ground decisions in verifiable evidence and mitigate hallucinations.
Despite its success in general reasoning tasks, Agentic RL remains underexplored for urban region profiling. 
Despite several pioneering efforts~\cite{liu2025citylens,urbanr1,liu2025cityrise}, existing methods still suffer from spurious causal reasoning.
In this work, we reformulate urban region profiling as a collaborative agentic task equipped with external tool calling~\cite{wang2025geovista}, enabling agents to dynamically query task-relevant external information and perform evidence-driven reasoning.

\subsubsection{Multi-Agent Systems (MAS)}

MAS leverages the collective intelligence of interacting agents to solve complex problems beyond the capacity of a single agent~\cite{shinn2023reflexion,schick2023toolformer,MACA,PMC,multiagent-survey,smilegeo,zheng2025graphgeo,agentsense}. Prior work shows that collaboration mechanisms—such as reflection~\cite{shinn2023reflexion}, tool use~\cite{schick2023toolformer,MACA}, planning~\cite{PMC}, and negotiation~\cite{multiagent-survey}—can significantly enhance problem-solving performance. Recent studies further demonstrate that Agentic RL enables scalable multi-agent decision-making through multi-turn environmental feedback~\cite{zhang2025agentrl,park2025maporl}.
In geospatial reasoning, graph-based multi-agent frameworks have been increasingly adopted to facilitate structured communication and consensus~\cite{smilegeo,hao2025unlockingsurvey}, highlighting the potential of collaborative reasoning for domain-intensive tasks. 
In this paper, we adopt a graph-based multi-agent collaborative reasoning framework to resolve the inconsistency across urban modalities.

\section{Methodology}

As shown in Figure~\ref{fig:framework}, \model~is a two-stage framework that integrates tool-augmented agentic learning with multi-agent collaborative reasoning for urban region profiling. 
In the first stage, modality-specific agents are trained via Supervised Fine-Tuning (SFT) cold start and Group Relative Policy Optimization (GRPO)-based reinforcement learning to actively invoke external tools and produce evidence-grounded predictions with confidence estimates. 
In the second stage, these agents interact through a structured collaboration graph with corroboration, rectification, and supervisory relations, enabling iterative conflict resolution and consensus consolidation beyond static multimodal fusion. 

\subsection{Tool-Augmented Agentic Learning}

In urban region profiling, the prediction of socio-economic indicators mainly depends on latent information that is not directly observable. As discussed in the Introduction, when relying solely on internal knowledge acquired through large-scale pretraining~\cite{geollm}, LLMs tend to exhibit inherent limitations related to correlation-driven inference. 
Moreover, urban analysis requires strong domain knowledge that is highly context-specific, which general-purpose LLMs largely lack~\cite{manvi2024large}. 
Therefore, \textbf{the incorporation of explicit tool chains and interpretable intermediate artifacts is necessary to suppress hallucinations and improve overall robustness}.
In this section, we describe the construction of the tool-use trajectory dataset and its use for training agentic RL.

\vspace{-0.5em}
\subsubsection{Tool-use Trajectory Curation Pipeline}
We select four representative categories of tools to construct the tool suite that are closely related to socio-economic indicators: (1) \textbf{Web Search}, (2) \textbf{Crop-and-Zoom}, (3) \textbf{Nightlight Retrieval} and (4) \textbf{Historical Satellite Imagery Retrieval}. See more details in Appendix~\ref{appendix:tooluse}.

Figure~\ref{fig:data_anno} illustrates the overall pipeline of Tool-use Trajectory Curation.
Given geo-referenced location samples with labels, the pipeline processes four complementary input modalities—satellite imagery, textual descriptions, POI distributions, and 3D building information. 
For different modalities, available tools are randomly sampled to drive the model to progressively acquire external evidence via multi-round interactions, while iteratively updating its judgments.
During trajectory curation, we employ Qwen3-VL-7B~\cite{Bai2025Qwen3VL} as the annotation MLLM, and enforce a unified output schema to record the complete sequence consisting of model reasoning, tool invocation, tool feedback, and iterative reasoning~\cite{react}.

\begin{figure}[t!]
  \centering
  \includegraphics[width=\linewidth]
  {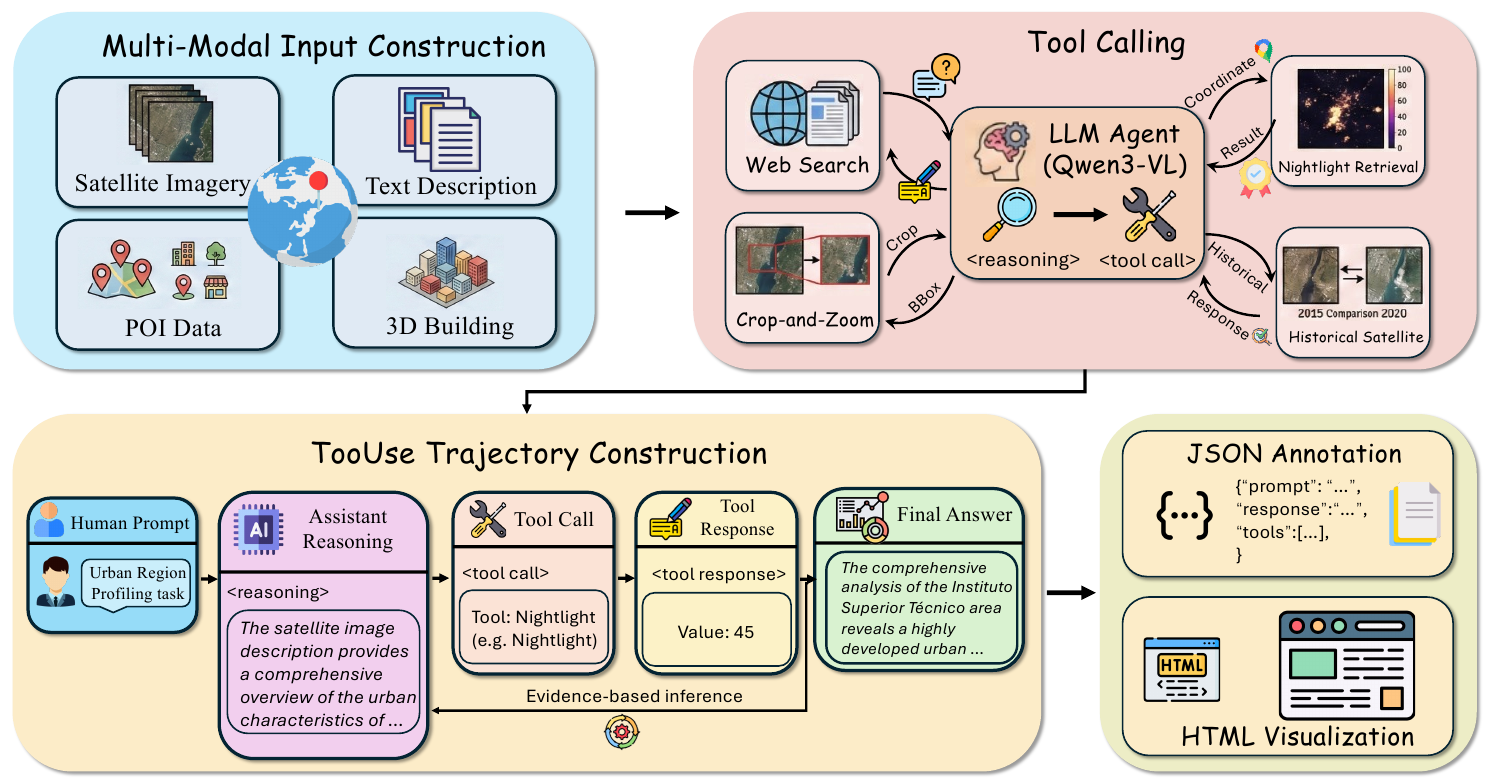}
  \vspace{-1em}
  \caption{Tool-use trajectory curation pipeline.}
  \vspace{-1em}
  \label{fig:data_anno}
\end{figure}

\subsubsection{Agentic Reinforcement Learning}

Recent urban foundation models~\cite{zhang2024urbansurvey,hao2025unlockingsurvey} have largely shifted from self-supervised pretraining~\cite{urbanvlp,urbanln,cao2026urbanmmcl} to instruction fine-tuning (SFT)~\cite{zhang2025urbanmllm,feng2025urbanllava}, achieving strong in-distribution performance but relying primarily on token-level supervision that fails to distinguish robust urban reasoning from spurious statistical correlations.
In contrast, reinforcement learning—particularly GRPO~\cite{guo2025deepseekr1,urbanr1,luo2025guir1,globe2025}—directly aligns optimization with verifiable, task-relevant rewards, enabling more stable and scalable learning of robust reasoning strategies.

In this work, we employ GRPO to incentivize tool-use reasoning capabilities in LLMs via reinforcement learning. 
In practice, tool invocation differs fundamentally from natural language dialogue, and models typically lack the necessary priors, such as reasoning-pattern priors (e.g., ReAct-style structures~\cite{react}), tool-timing priors (i.e., when to invoke tools), and tool-call formatting priors. 
Therefore, we first apply instruction-based supervised fine-tuning (SFT) to activate and ground the model’s tool-calling syntax, providing a stable foundation for subsequent RL optimization.
The core role of SFT cold start is to inject executable tool-use priors into agentic models, as pure reinforcement learning without such initialization often leads to unstable training dynamics and degenerate solutions~\cite{wu2024toolplanner,shabbir2025thinkgeo}.
The core idea of GRPO is to sample multiple candidate trajectories for a given input and update the policy based on relative feedback within each group, thereby eliminating the need for an absolute reward scale and providing more stable learning signals under heterogeneous reward functions~\cite{guo2025deepseekr1}.

\textbf{Tool-aware Composite Reward.}
For an input $x$ and a sampled response trajectory $y$, we define the total reward of three items:
\begin{equation}
  R_{\text{total}}(x,y)=\alpha\,R_{\text{outcome}}(x,y)+\beta\,R_{\text{format}}(x,y)+\gamma\,R_{\text{process}}(x,y),
\end{equation}
where $R_\text{outcome}$ measures predictive correctness (e.g., numeric accuracy), $R_\text{format}$ enforces strict compliance with the tool-calling / output grammar, and
$R_\text{process}$ encourages well-structured and appropriate tool use, preventing degeneration into tool-free, redundant, or invalid invocation behaviors (e.g., exceeding call limits or applying image-specific tools to text-only inputs).

\textbf{Group-wise Sampling and Relative Advantage}.
For each training instance $x$, we sample a group of $G$ candidates from the current policy $\pi_\theta$ and obtain their rewards $\{r_i\}_{i=1}^G$.
GRPO converts these absolute rewards into a group-relative advantage, which measures the relative performance of each candidate within the group.
Specifically, the advantage of candidate $i$ is defined as
\begin{equation}
A_i
= \frac{r_i - \frac{1}{G}\sum_{j=1}^{G} r_j}
{\sqrt{\frac{1}{G}\sum_{j=1}^{G}\left(r_j - \frac{1}{G}\sum_{k=1}^{G} r_k\right)^2}}.
\end{equation}

\textbf{Policy Optimization with KL Regularization.}
We update $\pi_\theta$ to increase the likelihood of candidates with positive relative advantage, while constraining drift from a reference policy $\pi_{ref}$
 via KL regularization:
 
\begin{equation}
\begin{aligned}
\mathcal{J}(\theta)
=&\;
\mathbb{E}_{\substack{
\{y_i\}_{i=1}^{G}\sim\pi_{\theta_{\text{old}}}(\cdot\mid x)}}
\Bigg[
\frac{1}{G}\sum_{i=1}^{G}
\min\!\Big(
\sigma_i A_i,\;
\mathrm{clip}(\sigma_i, 1-\epsilon, 1+\epsilon)\,A_i
\Big) \\
&\qquad\qquad
-\lambda\,\mathrm{KL}\!\left(
\pi_\theta(\cdot\mid x)\,\|\,\pi_{\text{ref}}(\cdot\mid x)
\right)
\Bigg],
\end{aligned}
\end{equation}

where $
\sigma_i = \frac{\pi_\theta(y_i\mid x)}{\pi_{\theta_{\text{old}}}(y_i\mid x)},
$
$\epsilon$ is the clipping threshold,
and $\lambda$ controls the weight of the KL regularization.
$\pi_{\text{ref}}$ denotes a fixed reference policy, initialized from the SFT model,
which constrains the policy update and preserves the base language behavior.

\subsection{Multi-Agent Collaborative Reasoning}
After obtaining initial predictions from modality-specific agents via tool-augmented agentic learning, our Multi-Agent Collaborative Reasoning proceeds in two stages: (1) Multi-Agent Interaction Graph construction and (2) Relation-Aware Iterative Reasoning.

\subsubsection{Multi-Agent Interaction Graph}

Multi-agent collaboration organizes agents associated with distinct modalities or perspectives into a cooperative system that enables structured interaction and cross-verification. 
This transforms multimodal fusion from static representational alignment into a dynamic reasoning process and improving the robustness and controllability of heterogeneous information fusion.

As illustrated in Figure~\ref{fig:framework} (b), we propose a heterogeneous graph-based framework for multi-agent collaborative reasoning.
In practice, multi-agent interactions are inherently heterogeneous, which we summarize into three categories: \ding{182} Consistent predictions call for confirmation and consolidation; \ding{183} Conflicting predictions require error rectification, and \ding{184} When agent confidence exhibits strong asymmetry, the interaction topology becomes supervision-oriented.
Motivated by this observation, we construct three types of edges and allow them to dynamically express three interaction semantics—\textbf{Corroboration, Rectification and Supervision}.

In the beginning, 
we employ
$K$ modality-specific, tool-augmented agents, denoted by
$\mathcal{V}=\{v_1,v_2,…,v_K\}$. 
Each agent $v_k$
produces indicator predictions $y_k$
and an associated confidence score $c_k \in [0, 1]$. 
The collaborative reasoning process can be modeled as a graph \( \mathcal{G} = (\mathcal{V}, \mathcal{E}) \) where nodes $\mathcal{V}$ represent agents and edges $\mathcal{E}$ encode their interaction patterns.
the edge set is defined as
$
  \mathcal{E}
  = \mathcal{E}^{\mathrm{cor}} \;\cup\; \mathcal{E}^{\mathrm{rec}} \;\cup\; \mathcal{E}^{\mathrm{sup}} .
$

Next, we introduce three kinds of edges in detail.

\textbf{(1) Corroboration Edge.}
Corroboration edges connect agents whose predictions are sufficiently close, indicating mutually supportive predictions that consolidate consensus and reinforce consistent interpretations.
A corroboration edge is constructed as
\begin{equation}
  (v_j \rightarrow v_i) \in \mathcal{E}^{\mathrm{cor}}
  \quad \text{if} \quad
  \left\|~\mathrm{pred}_i - \mathrm{pred}_j\right\| < \tau_{\mathrm{cor}},
\end{equation}
where $\tau_{\mathrm{cor}}$ is corroboration threshold.
Intuitively, this relation corresponds to mutual confirmation and evidence reinforcement, amplifying reliable shared information and stabilizing consensus.

\textbf{(2) Rectification Edge.}
Rectification edges explicitly model conflict-aware correction.
When agent $v_i$’s hypothesis shows significant inconsistency with other agents, other agents are encouraged to challenge and correct it.
A rectification edge is constructed as
\begin{equation}
  (v_j \rightarrow v_i) \in \mathcal{E}^{\mathrm{rec}}
  \quad \text{if} \quad
  \left\|~\mathrm{pred}_i - \mathrm{pred}_j\right\| > \tau_{\mathrm{rec}},
\end{equation}
where $\tau_{\mathrm{rec}}$ is rectification threshold.
This relation explicitly exposes cross-modal inconsistency and drives hypothesis refinement.

\textbf{(3) Supervisory Edge.}
While corroboration and rectification explicitly address agreement and conflict among agents, they do not determine which agent should exert greater influence during the reasoning process. At this stage, the confidence estimation mechanism plays a critical role by guiding the direction and strength of inter-agent interactions.
When the confidence difference between two agents' predictions exceeds a threshold $\tau_{\mathrm{sup}}$, we construct a directed supervisory edge as
\begin{equation}
  (v_j \rightarrow v_i) \in \mathcal{E}^{\mathrm{sup}}
  \quad \text{if} \quad
  c_j - c_i > \tau_{\mathrm{sup}}.
\end{equation}
This structure forms a guidance-oriented topology analogous to teacher-forcing knowledge distillation, enabling high-confidence agents to exert stronger influence.

\subsubsection{Relation-Aware Iterative Reasoning}
After the construction of multi-agent interaction graph, agents iteratively refine their predictions through multi-round interactions. 
Each agent maintains a hidden state initialized from its modality-specific encoding and initial predictions.
\begin{equation}
  \mathbf{h}_i^{(0)} =
  \mathrm{Init}_{\theta}~\!\Big(\mathrm{pred}_i,\; \mathrm{Embed}_{\psi}(\mathbf{s}_i),\; c_i\Big),
\end{equation}
where $\mathbf{h}_i^{(0)} \in \mathbb{R}^{d_h}$ denotes the initial hidden state of agent $i$, $\mathrm{pred}_i$ is its initial indicator prediction, $\mathbf{s}_i$ is the modality-specific reasoning text, $c_i \in [0,1]$ is the associated confidence score. $\mathrm{Embed}_{\psi}(\cdot)$ maps $\mathbf{s}_i$ into a shared embedding space using an LLM-based encoder, and $\mathrm{Init}_{\theta}(\cdot)$ is a learnable initialization function (e.g., MLP layers) that fuses these information to produce $\mathbf{h}_i^{(0)}$.

Subsequently, the agents engage in $L$ rounds of iterative interaction.
At each interaction round $\ell$,
for each agent $v_i$, we partition its neighborhood $\mathcal{N}(i)$ by edge type:
\begin{equation}
  \mathcal{N}^{t}(i)=\{\,v_j \mid (v_j \!\rightarrow\! v_i)\in \mathcal{E}^{t}\,\}, \quad t\in\mathcal{R},
\end{equation}
where $\mathcal{R}=\{\mathrm{cor},\,\mathrm{rec},\,\mathrm{sup}\}$. The node-level interaction then produces relation-specific messages, enabling distinct message-passing mechanisms tailored to corroboration, rectification, and supervision influence:

\begin{itemize}[label=$\diamond$,leftmargin=*]
  \item
    Corroboration Message
    \begin{equation}
      \begin{gathered}
        g_{ji}
        = \sigma~\!\Big(
          W_{\mathrm{cor}}\,[h_j^{(l)} \,\|\, h_i^{(l)}]
          + b_{\mathrm{conf}}\,(c_{v_j}-c_{v_i})
        \Big),\\
        \mu_{ji}^{\mathrm{cor}}
        = g_{ji}\odot (W_g\,h_j^{(l)}),
      \end{gathered}
    \end{equation}
    where $W_{\mathrm{cor}}$ denotes a learnable weight matrix,
$b_{\mathrm{conf}}$ is learnable that modulates the influence of
confidence differences,
$\sigma(\cdot)$ denotes sigmoid activation,
and $\mu_{ji}^{\mathrm{cor}}$ represents the corroboration message
propagated from node $v_j$ to node $v_i$.
    
  \item
    Rectification Message
    \begin{equation}
      \mu_{ji}^{\mathrm{rec}}
      = W_{\mathrm{rec}} \cdot \text{ReLU}(h_j^{(l)} - h_i^{(l)}),
    \end{equation}
where $W_{\mathrm{rec}}$ denotes a learnable transformation matrix,
and $\mu_{ji}^{\mathrm{rec}}$ represents the rectification message
propagated from $v_j$ to $v_i$.

  \item
    Supervisory Message
    \begin{equation}
    \nonumber
      \begin{gathered}
        \alpha_{ji}
=
\frac{
\exp\!\left(
c_j \, {h_j^{(l)}}^\top W_{\mathrm{attn}} \, h_i^{(l)}
\right)
}{
\sum\limits_{z \in \mathcal{N}^{\mathrm{sup}}(i)}
\exp\!\left(
c_z \, {h_z^{(l)}}^\top W_{\mathrm{attn}} \, h_i^{(l)}
\right)
}, \quad \mu_{ji}^{\mathrm{sup}}
        = \alpha_{ji}\, W_{\mathrm{sup}}(h_j^{(l)}),
        \end{gathered}
    \end{equation}
where $W_{\mathrm{attn}}$ denotes a learnable attention weight matrix,
$W_{\mathrm{sup}}$ is a learnable transformation matrix for supervisory
message construction, and $\mu_{ji}^{\mathrm{sup}}$ represents the
supervisory message propagated from node $v_j$ to node $v_i$.
\end{itemize}

The aggregated message for each agent is computed via relation-aware aggregation.
\begin{equation}
m_i^{t}
=
\frac{1}{\text{max}(1, \lvert \mathcal{N}^{t}(i) \rvert) }
\sum_{j \in \mathcal{N}^{t}(i)}
\mu_{ji}^{t},
\quad
\mu_i
=
W_f
\big[
m_i^{\mathrm{cor}}
\;\|\;
m_i^{\mathrm{rec}}
\;\|\;
m_i^{\mathrm{sup}}
\big],
\end{equation}
where $t\in 
\{\mathrm{cor},\,\mathrm{rec},\,\mathrm{sup}\}$,
$\mathbf{W}_{f}$ represents learnable parameters.
The agent state is then updated via a gated update function.

\begin{equation}
h_i^{(l+1)} = \mathrm{GRU}(h_i^{(l)}, \mu_i^{(l)})
\end{equation}
After each round, the agent confidence is recalibrated as $c_i^{(l)}=\sigma~(\mathbf{w}^\top\mathbf{h}_i^{(l)}+b)$.

\begin{itemize}[label=$\diamond$,leftmargin=*]
\item Hierarchical Final Judge.
After
$L$ interaction rounds, each agent yields a calibrated hidden state and a confidence sequence.
To avoid relying solely on the final-round output, we introduce a Hierarchical Spatio-Temporal Judge to jointly evaluate process reliability and cross-agent contribution.

The temporal judge evaluates each agent’s calibration trajectory, capturing stability and uncertainty across rounds.
\begin{equation}
  \tilde{h}_i =
  \mathrm{TemporalEnc}(h_i^{(1:L)}, c_i^{(1:L)}).
\end{equation}


The spatial judge performs cross-agent attention over temporally refined representations to compute importance weights and produce the final prediction.
\begin{equation}
\hat{Y}
=
\sum_{i=1}^{K}
\frac{\exp\!\big(\mathrm{Attn}(\tilde{h}_i)\big)}
{\sum_{j=1}^{K} \exp\!\big(\mathrm{Attn}(\tilde{h}_j)\big)}
\cdot g(\tilde{h}_i),
\end{equation}
where $g(\cdot)$ denotes learnable prediction head.
\end{itemize}

\begin{table*}[t]
  \centering
    \caption{Performance comparison under In-Domain and Unseen Cities settings. AgentBlender\_w denotes confidence 
weighted mean, while AgentBlender\_v stands for majority voting. UrbanVLP* represents UrbanVLP without street-view imagery branch.}
\vspace{-1em}
  \setlength{\tabcolsep}{6pt}
  \renewcommand{\arraystretch}{0.85}
  \resizebox{\textwidth}{!}{%
    \tiny
    \begin{tabular}{l|cc|cc|cc|cc|cc|cc}
      \toprule
      \multirow{3}{*}{\textbf{Model}} &
      \multicolumn{6}{c|}{\textbf{In-Domain}} &
      \multicolumn{6}{c}{\textbf{Unseen Cities}} \\
      \cmidrule(lr){2-7}\cmidrule(lr){8-13}
      & \multicolumn{2}{c|}{\textbf{Carbon}} &
      \multicolumn{2}{c|}{\textbf{Population}} &
      \multicolumn{2}{c|}{\textbf{GDP}} &
      \multicolumn{2}{c|}{\textbf{Carbon}} &
      \multicolumn{2}{c|}{\textbf{Population}} &
      \multicolumn{2}{c}{\textbf{GDP}} \\
      \cmidrule(lr){2-3}\cmidrule(lr){4-5}\cmidrule(lr){6-7}
      \cmidrule(lr){8-9}\cmidrule(lr){10-11}\cmidrule(lr){12-13}
      & $R^2$ & $\rho$ & $R^2$ & $\rho$ & $R^2$ & $\rho$
      & $R^2$ & $\rho$ & $R^2$ & $\rho$ & $R^2$ & $\rho$ \\
      \midrule

      Qwen2.5-VL-3B          &  &  &  &  &  &  &  &  &  &  &  &  \\
      \quad\quad\quad\quad ——Satellite        & 0.164  & 0.541 & 0.153 & 0.552 & 0.194  & 0.582 & 0.136 & 0.561 & 0.152 & 0.587 & 0.203 & 0.614 \\
      \quad\quad\quad\quad ——Text             & 0.175 & 0.457 & 0.129 & 0.518 & 0.303 & 0.564 & 0.262 & 0.593 & 0.153 & 0.553 & 0.251 & 0.601 \\
      \quad\quad\quad\quad ——POI              & 0.159 & 0.482 & 0.109 & 0.447 & 0.145 & 0.474 & 0.132 & 0.457 & 0.126 & 0.498 & 0.108 & 0.373 \\
      \quad\quad\quad\quad ——3D               & 0.172 & 0.542 & 0.146 & 0.494 & 0.228 & 0.600 & 0.178 & 0.598 & 0.247 & 0.574 &0.188  & 0.608 \\
      Qwen3-VL-4B            &  &  &  &  &  &  &  &  &  &  &  &  \\
      \quad\quad\quad\quad ——Satellite        & 0.321 & 0.684 & 0.183 & 0.562 & 0.284 &0.667  & 0.289& 0.608& 0.204 & 0.535 & 0.262 & 0.564 \\
      \quad\quad\quad\quad ——Text             &0.314  & 0.581 & 0.143 & 0.546 &0.311 & 0.640 &  0.319& 0.619 & 0.237 & 0.598 & 0.247 & 0.627 \\
      \quad\quad\quad\quad ——POI              & 0.265 & 0.556 & 0.259 & 0.635 & 0.235 & 0.554 & 0.285 & 0.627 & 0.224 & 0.648 & 0.215 & 0.582 \\
      \quad\quad\quad\quad ——3D               & 0.201 & 0.579 & 0.167 & 0.569 & 0.248 & 0.578  & 0.202 & 0.564 & 0.085 & 0.709 & 0.299 & 0.588 \\
      
      InternVL2.5-4B  &  &  &  &  &  &  &  &  &  &  &  &  \\
      \quad\quad\quad\quad ——Satellite        & 0.089 & 0.375 & -0.120 & 0.222 & 0.056 & 0.327 & 0.132 & 0.473 & -0.040 & 0.408 & 0.032 & 0.371 \\
      \quad\quad\quad\quad ——Text             & 0.186 & 0.474 & 0.226 & 0.513 & 0.181 & 0.538 & 0.159 & 0.455 & 0.219 & 0.520 & 0.120 & 0.466 \\
      \quad\quad\quad\quad ——POI              & -0.159 & 0.282 & -0.139 & 0.347 & 0.145 & 0.474 & -0.133 & 0.216 & -0.202 & 0.198 & -0.081 & 0.307 \\
      \quad\quad\quad\quad ——3D               & 0.143 & 0.542 & 0.126 & 0.494 & 0.228 & 0.544 &0.158  & 0.538 & 0.124 & 0.466 & -0.018 & 0.612 \\
      
      Kimi-VL-A3B        &   &  &  &  &  &&&&&&  \\
      \quad\quad\quad\quad ——Satellite        &   0.065 & 0.381 & -0.179 & 0.320 & 0.122 & 0.417 & 0.042 & 0.289 & 0.049 & 0.445 & 0.083 & 0.545 \\
      \quad\quad\quad\quad ——Text             & -0.109 & 0.532 & 0.005 & 0.538 & 0.078 & 0.561 & -0.176 & 0.496 & 0.092 & 0.633 & -0.098 & 0.527 \\
      \quad\quad\quad\quad ——POI              & 0.121 & 0.359 & 0.208 & 0.437 & 0.153 & 0.503 & -0.171 & 0.033 & -0.085 & 0.142 & -0.098 & 0.251 \\
      \quad\quad\quad\quad ——3D               & 0.243 & 0.542 & 0.227 & 0.494 & 0.328 & 0.612 & 0.178 & 0.598 & 0.210 & 0.566 & -0.019 & 0.608 \\
      \midrule
      GPT-4o                 &  &  &  &  &  &  &  &  &  &  &  &  \\
      \quad\quad\quad\quad ——Satellite        & 0.460 & 0.770 & 0.431 & 0.652 & 0.520 & \textbf{0.816} & 0.336 & 0.752 & 0.477 & 0.709 & 0.138 & 0.725 \\
      \quad\quad\quad\quad ——Text             & 0.344 & 0.643 & 0.231 & 0.549 & 0.388 & 0.651 & 0.296 & 0.653 &0.201  &0.492  & 0.303 & 0.721 \\
      \quad\quad\quad\quad ——POI              & 0.326 &0.740  &0.419  &0.617  & 0.353 &0.634  & 0.327 & 0.733 & 0.411 & 0.715 & 0.318 &0.702  \\
      \quad\quad\quad\quad ——3D               &  0.559& 0.767 & \underline{0.554} & \underline{0.783} & 0.432 &0.652  & 0.522 & 0.748 & 0.589 & 0.755 & 0.385 & 0.734 \\
      
      Gemini-2.5-Flash         &  &  &  &  &  &  &  &  &  &  &  &  \\
      \quad\quad\quad\quad ——Satellite        & 0.533 &0.768  & 0.532 & 0.801 & 0.381 &0.639  & 0.496 & 0.744 & \underline{0.502} & \underline{0.751} & 0.358 & \underline{0.757} \\
      \quad\quad\quad\quad ——Text             &0.421  & 0.693 & 0.226 & 0.603 & 0.406 & 0.674 & 0.388 & 0.679 &  0.259& 0.623 & 0.386 &0.673  \\
      \quad\quad\quad\quad ——POI              &  0.414& 0.725 & 0.337 & 0.635 & 0.408 & 0.765 & 0.337& 0.742& 0.339& 0.648& 0.266&0.722 \\
      \quad\quad\quad\quad ——3D               & 0.535 & 0.753 & 0.238 & 0.573 & 0.515 & 0.754 & 0.517 & 0.759 & 0.268 & 0.583 & 0.365 &0.671  \\
      
      Ours w/o MAS         &  &  &  &  &  &  &  &  &  &  &  &  \\
      \quad\quad\quad\quad ——Satellite        & 0.386 & 0.677 & 0.255 & 0.522 & 0.387 & 0.682 & 0.369 & 0.656 & 0.250 & 0.543 & 0.237 &0.630  \\
      \quad\quad\quad\quad ——Text             & 0.428 & 0.676 & 0.220 & 0.574 & 0.443 & 0.671 & 0.453 & 0.625 & 0.221 &0.550  &0.416  &0.670  \\
      \quad\quad\quad\quad ——POI              & 0.329 & 0.586 & 0.291 & 0.568 & 0.285 & 0.605 &  0.326& 0.594 &0.120  & 0.575 & 0.322 & 0.646 \\
      \quad\quad\quad\quad ——3D               & 0.365 & 0.664 & 0.207 & 0.573 & 0.416 & 0.694 & 0.354 & 0.669 & 0.204 & 0.552 & 0.429 & 0.763 \\

      \midrule
            UrbanCLIP              & 0.544 & 0.743 & 0.445 & 0.620 & 0.532 & 0.732 & 0.421 & 0.690 & 0.382 & 0.654 &0.522  &0.751  \\
      UrbanVLP*               & 0.551 & 0.747 & 0.448 & 0.631 & 0.526 & 0.728 & 0.430 & 0.682 & 0.383 & 0.657 & 0.518 & 0.742 \\
      \midrule
      AgentFusion              & 0.512 & 0.721 & 0.383 & 0.645  & 0.551 & 0.748 &0.472 & 0.704& 0.409 & 0.676 & \underline{0.536} & 0.738 \\
      AgentBlender\_w              & 0.504 & 0.718 & 0.378 & 0.647 & 0.543 & 0.745 & 0.516 & 0.713 & 0.308 & 0.629 & 0.532 & 0.782 \\
      AgentBlender\_v               &  0.441& 0.678 & 0.262 & 0.585 & 0.535 & 0.735 & 0.429 & 0.638 & 0.310 & 0.614 & 0.508 & 0.708 \\
      Agent-Arbiter    & \underline{0.564} & \underline{0.771} & 0.355 & 0.725 & \underline{0.553} & \underline{0.772} & \underline{0.530} & \underline{0.724} & 0.345 & 0.709 & 0.521 & 0.715 \\
      Hierarchical Prompting               & 0.311 & 0.606 & 0.215 & 0.542 &0.194  & 0.556 & 0.279 & 0.547 & 0.261 & 0.572 & 0.147 & 0.487 \\          
      Ours                   &  \textbf{0.638}& \textbf{0.800} & \textbf{0.649} & \textbf{0.803} & \textbf{0.626} & 0.799  & \textbf{0.608} & \textbf{0.768} & \textbf{0.621} & \textbf{0.779} &\textbf{0.598}&\textbf{0.786}  \\
      \bottomrule
    \end{tabular}%
  }

\vspace{-1em}
  \label{tab:main_results}
\end{table*}

\section{Experiments}
In this section, we conduct extensive experiments to investigate
the following Research Questions (RQ):
\begin{itemize}
    [leftmargin=*]
    \item \textbf{RQ1}: Does \model~achieve superior performance over state-of-the-art DL-based and multi-agent baselines across multiple urban socio-economic prediction tasks?
    \item \textbf{RQ2}: What are the individual contributions of the various components of \model~to its overall effectiveness ?
    \item \textbf{RQ3}: How well does \model~generalize to unseen cities ?
    \item \textbf{RQ4}: How does \model~perform qualitatively in practice?
\end{itemize}

\subsection{Experimental Setup}
\subsubsection{Datasets}
In this work, we construct our dataset from four kinds of representative modalities: \textit{Satellite Imagery, Text, POI, 3D}.
Following~\cite{urbanclip}, we evaluate \model~on three classical socio-economic indicators: \textit{Carbon emissions, GDP}, and \textit{Population}. Following~\cite{geollm,urbanr1}, we sample locations globally and provide the model with longitude–latitude coordinates, addresses, and nearby places. 
We collect 2,000 locations in total, each covering a 1km×1km area,  and split them into training, validation, and test sets with a ratio of 7:1:2.
Following the practice of~\cite{geollm}, we convert raw indicator values into relative magnitude scores and scale them to the range [0, 9.9], enabling stable learning across heterogeneous socio-economic variables with disparate units and scales.
In total, we curate 300 × 3 cold-start reasoning trajectories for SFT and 1000 × 3 samples for RL, each corresponding to a specific region–indicator pair.
More Details about dataset are provided in Appendix~\ref{appendix:datasource}.

Figure~\ref{fig:data_stat} illustrates the statistics of our dataset.
Figure~\ref{fig:data_stat} (a) presents the global distributions of the training and test sets.
Figure~\ref{fig:data_stat} (b) visualizes the meta information of the training set, including addresses and nearby places, in the form of a word cloud.
Figure~\ref{fig:data_stat} (c) shows the statistical breakdown of the POI data using a pie chart.

\begin{figure}[ht]
  \centering
  \includegraphics[width=\linewidth]
  {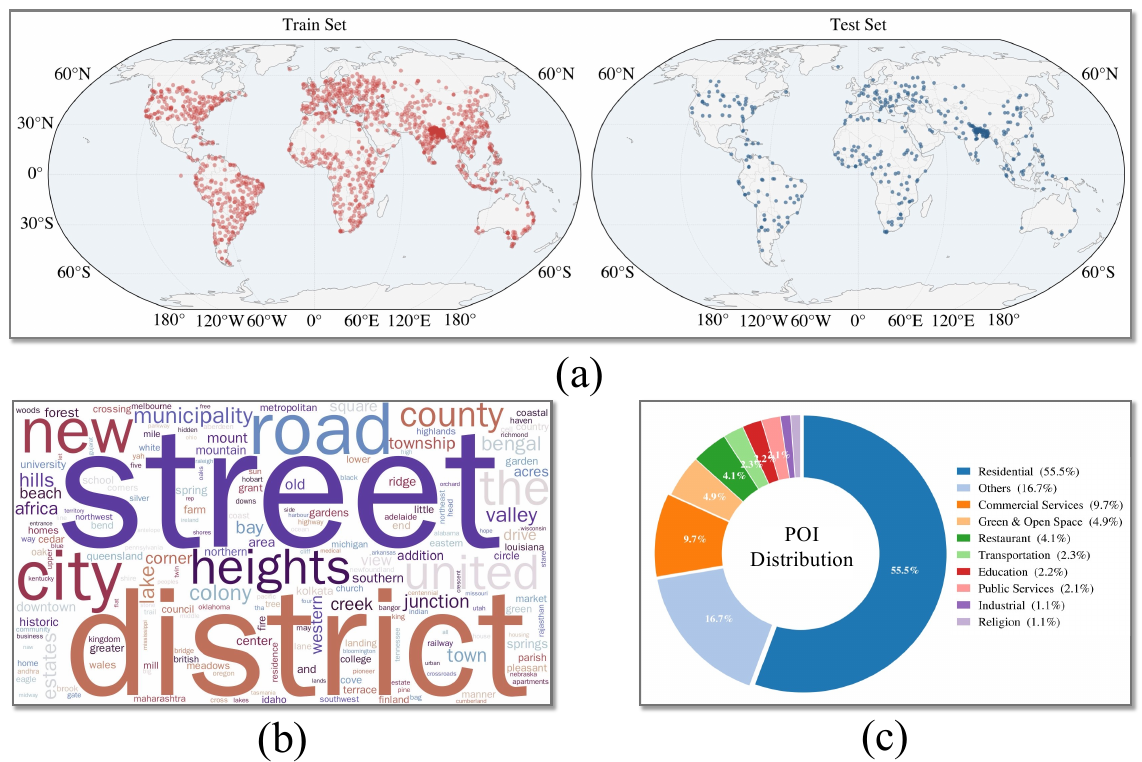}
  \vspace{-2.2em}
  \caption{Dataset Analysis.}
  \label{fig:data_stat}
  \vspace{-1.7em}
\end{figure}

\subsubsection{Baseline}
We compare \model~with the following baselines on our dataset: Open-source LLMs like Qwen2.5-VL-3B~\cite{bai2025qwen25vl}, Qwen3-VL-4B~\cite{Bai2025Qwen3VL}, InternVL2.5-4B~\cite{internvl25} and Kimi-VL-A3B~\cite{team2025kimivl}; close-source LLMs like GPT-4o~\cite{hurst2024gpt4o} and Gemini-2.5-Flash~\cite{comanici2025gemini25}; Conventional pretraining methods like UrbanCLIP~\cite{urbanclip} and UrbanVLP~\cite{urbanvlp}.
We also compared different multi-agent collaboration methods like AgentFusion, AgentBlender, Agent-Arbiter and Hierarchical Prompting.
Please refer to the Appendix~\ref{appendix:baseline} for more details.

\subsubsection{Evaluation Metrics and Implementation}

We evaluate model performance using the coefficient of determination ($R^2$) and Spearman’s rank correlation coefficient ($\rho$). $R^2$ measures how much variance in the ground-truth values is explained by the predictions, while Spearman’s $\rho$ evaluates the monotonicity of the relationship, capturing the model's ability to rank socio-economic indicators correctly irrespective of linear scale. 

All experiments are conducted on A800 GPUs using Qwen3-VL-4B~\cite{Bai2025Qwen3VL} as the backbone for training tool-use capabilities and multi-agent collaborative reasoning. 
Multi-agent collaborative reasoning is implemented with a relation-aware debate mechanism, where agent interactions are dynamically governed by three thresholds: 
$\tau_{cor}$=0.4, $\tau_{rec}$=1.2, $\tau_{sup}$=0.25, enabling balanced consensus reinforcement, conflict rectification, and confidence-guided supervision. 
The number of interaction rounds $L$ is set to 5. $d_h$ is set to 192.
The training pipeline is built on VeRL, with a batch size of 8, a learning rate of 1e-6, and a maximum of 4 tool-use turns per trajectory to control reasoning depth and computational cost.
In Multi-Agent Collaborative Reasoning, the batch size is set to 128, with a learning rate of 1e-4. The model is optimized by AdamW~\cite{adamw}.

\subsection{RQ1: Overall Performance}

To demonstrate the effectiveness of \model, we compare it against state-of-the-art models, including representative open-source, closed-source, and multi-agent systems.
Table~\ref{tab:main_results} summarizes the overall performance of all methods evaluated in our experiments, from which we can obtain the following findings: 
\textbf{i) \model~consistently achieves superior performance across all socio economic prediction tasks.}
Compared with the strongest baselines, \model~ demonstrates notable improvements on Carbon Emission, Population, and GDP, with 
$R^2$ gains of 7.4\%, 9.5\%, and 7.3\%, respectively, while preserving high Spearman correlation. This suggests that \model~improves both predictive accuracy and the stability of regional ranking.
These results highlight the advantage of explicit multi-agent collaboration over DL-based static aggregation or single-decision paradigms.
\textbf{ii) The performance gains vary across different indicators, reflecting heterogeneous modality contributions.}
\model~achieves the strongest performance on Population, followed by Carbon Emission, while GDP remains relatively more challenging. This trend suggests that indicators closely tied to physical urban form and functional distribution (e.g., residential density and land use) benefit more from collaborative reasoning, whereas purely economic indicators rely on more implicit and noisy signals.

In comparison, closed-source models generally outperform open-source models.
Static fusion approaches (e.g., AgentFusion and AgentBlender) exhibit capability in complex urban prediction tasks. Centralized arbitration methods (e.g., Agent-Arbiter) can further improve performance but remain constrained by single-point decision-making. In contrast, sequential hierarchical prompting introduces structured reasoning but is susceptible to error accumulation.
These differences suggest that the effectiveness of multi-agent systems depends not only on the inclusion of multiple specialists, but more critically on the design of their collaboration and information interaction mechanisms.

\begin{figure}[!b]
  \centering
  \begin{subfigure}[t]{\linewidth}
    \centering
    \includegraphics[width=\linewidth]{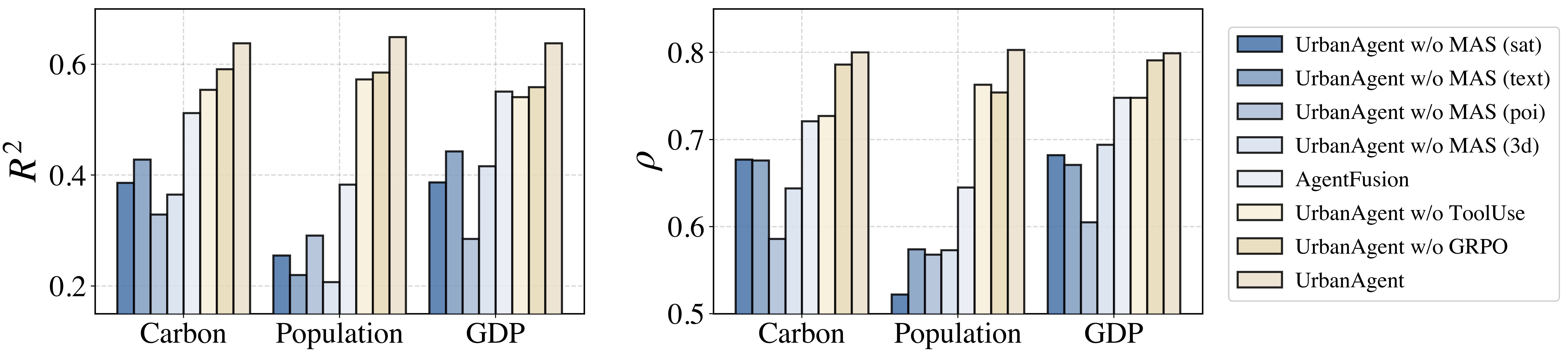}
    \vspace{-1em}
    \caption{Ablation results on $R^2$ and $\rho$.}
    \label{fig:ablation}
  \end{subfigure}
  \vspace{1em}
  \begin{subfigure}[t]{\linewidth}
    \centering
    \includegraphics[width=\linewidth]{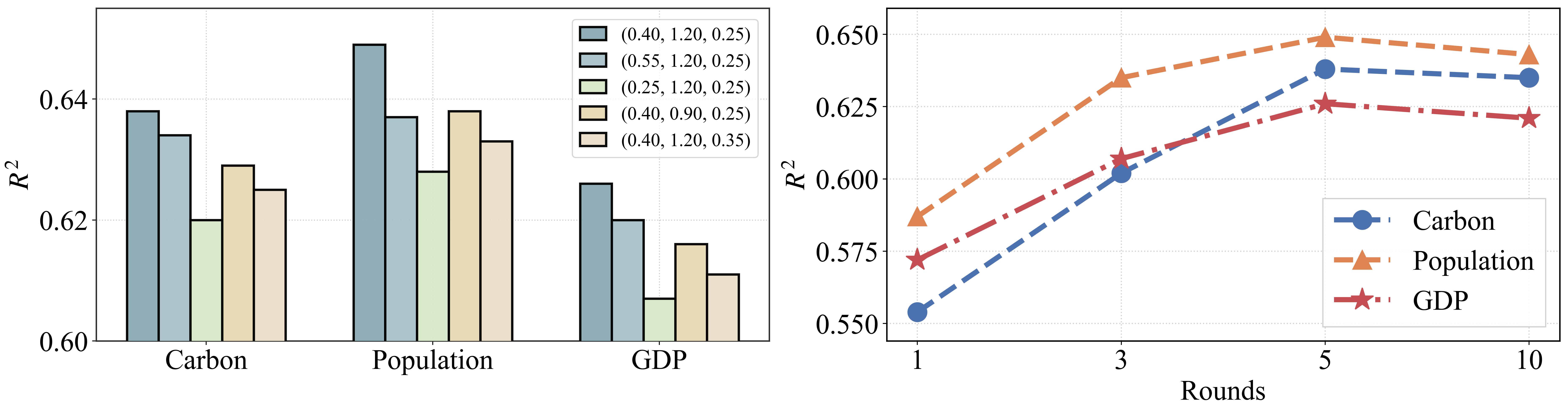}
    \vspace{-1em}
    \caption{Hyperparameter Analysis.}
    \label{fig:hyper_param}
  \end{subfigure}
  \vspace{-1em}
  \caption{Ablation and hyperparameter analysis.}
  \label{fig:ablation_and_hyper}
\end{figure}

\subsection{RQ2: Ablation Studies}
We conduct ablation studies to assess the impact of individual components in \model~(Figure~\ref{fig:ablation}). Removing the multi-agent collaboration mechanism (\model~w/o MAS) leads to consistent performance degradation across all tasks, particularly for \textit{Population} prediction, underscoring the importance of inter-agent interaction for resolving cross-modal inconsistencies.
Disabling tool usage results in an average performance drop of about 13\%, which is less severe than that of \model~w/o MAS, showing that multi-agent collaboration contributes more substantially to overall performance. Moreover, the degradation observed when tools are used without GRPO suggests that SFT alone, while enabling instruction following, is insufficient to bridge the accuracy gap.

We further analyze sensitivity to key hyperparameters in Figure~\ref{fig:hyper_param}. Different configurations yield noticeable performance variations, with ($\tau_{cor}$, $\tau_{rec}$, $\tau_{sup}$)=(0.40, 1.20, 0.25) achieving the best overall results, reflecting a balanced trade-off between agent agreement and conflict resolution. Increasing the number of reasoning rounds improves performance up to five rounds, beyond which gains plateau or slightly decline, indicating limited performance gains from additional interaction rounds.

\subsection{RQ3: Transferability Study}
To evaluate out-of-distribution generalization, we conduct experiments under the Unseen Cities setting, where test samples are drawn from cities absent during training. The evaluation set includes 200 locations from four countries—Japan, Norway, Madagascar, and Mexico—covering diverse geographic and economic conditions. This setting requires models to generalize beyond city-specific statistics and rely on transferable urban cues.

As shown in Table~\ref{tab:main_results}, \model~consistently achieves the best performance across all indicators and effectively transfers to unseen urban environments. While some baselines show occasional gains for certain indicators, these improvements are neither consistent across tasks nor stable across metrics, and often reflect reduced data complexity rather than genuine transferability. In contrast, \model~exhibits uniformly small and stable performance degradation, indicating robust, city-agnostic reasoning rather than reliance on distributional artifacts.

\subsubsection{Illustration of Tool-Use Example}

As shown in Table~\ref{tab:main_results}, the comparison between Ours w/o MAS and Qwen3-VL-4B highlights the effectiveness of tool-use in enhancing reasoning accuracy and prediction reliability. To illustrate this more intuitively, we present a representative example in Figure~\ref{fig:tooluse}.
The model first conducts a preliminary morphological analysis from a satellite image to form initial socio-economic analysis based on visual cues such as building density and road connectivity. It then actively identifies missing information and invokes the \textit{web\_search} tool to incorporate external socio-economic facts about Longhua District, grounding its assessment beyond static visual correlations.

\subsection{RQ4: Qualitative Analysis}

\begin{figure}[!b]
  \centering
  \vspace{-0.5em}
  \includegraphics[width=\linewidth]
  {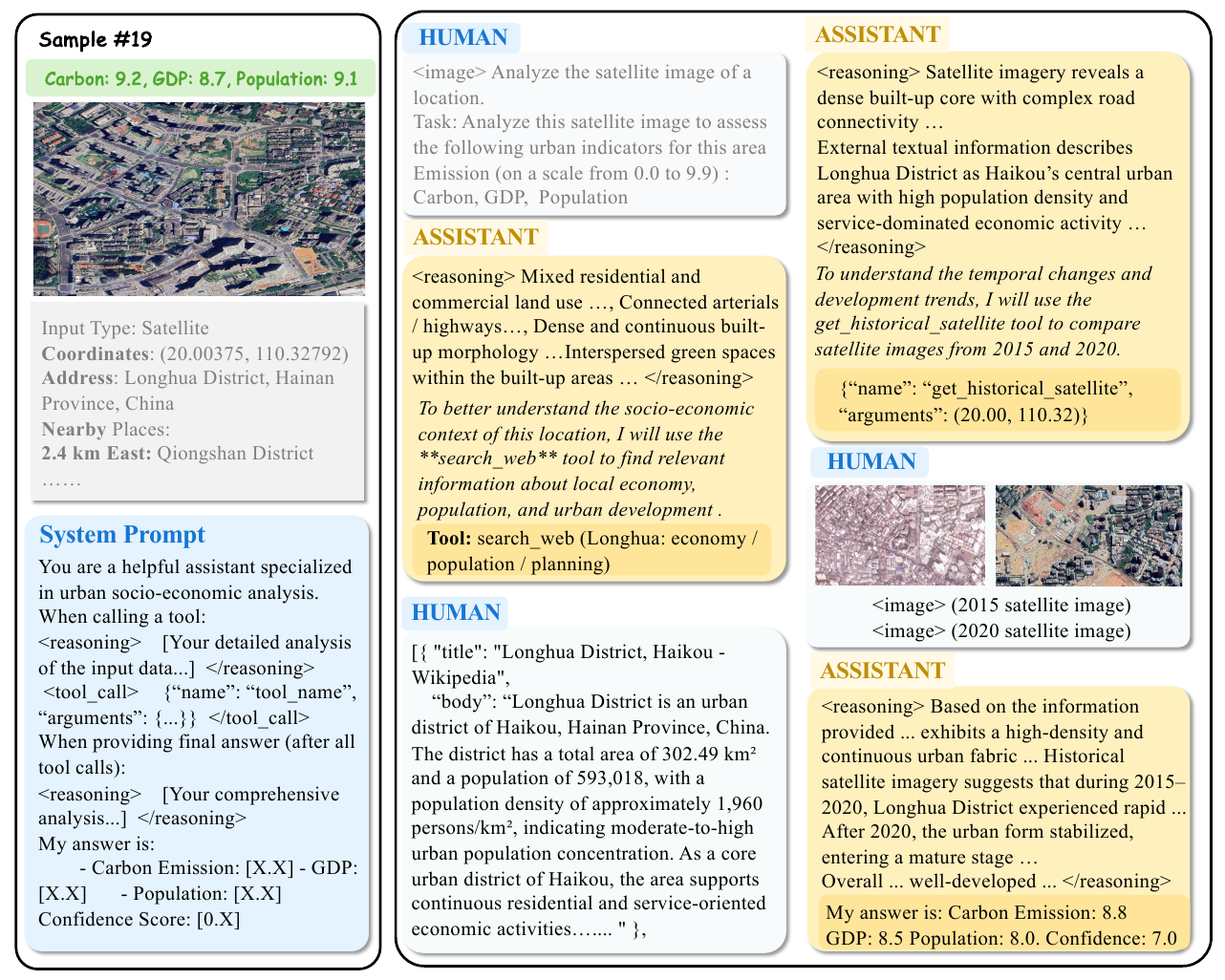}
  \vspace{-1.5em}
  \caption{Tool-Use Example Illustration. }
  \label{fig:tooluse}
\end{figure}

The web search results supplement information on population and economic conditions.
To further disambiguate between a mature urban core and a rapidly expanding area (the former typically exhibiting higher and more stable socio-economic indicators), the model calls the \textit{historical satellite} tool and analyzes imagery from 2015 to 2020, revealing a trajectory of rapid development followed by stabilization. Through iterative reasoning with tool invocation and evidence feedback, the model refines its predictions and produces estimates consistent with the region’s development stage and associated with high confidence.
Due to space constraints, some information is omitted in Figure~\ref{fig:tooluse}. Please refer to Figure~\ref{fig:detail_tooluse_example} in the Appendix for a detailed illustration.

\subsubsection{Case Study of Multi-Agent Collaborative Reasoning}
\begin{figure}[!t]
  \centering
  \includegraphics[width=\linewidth]
  {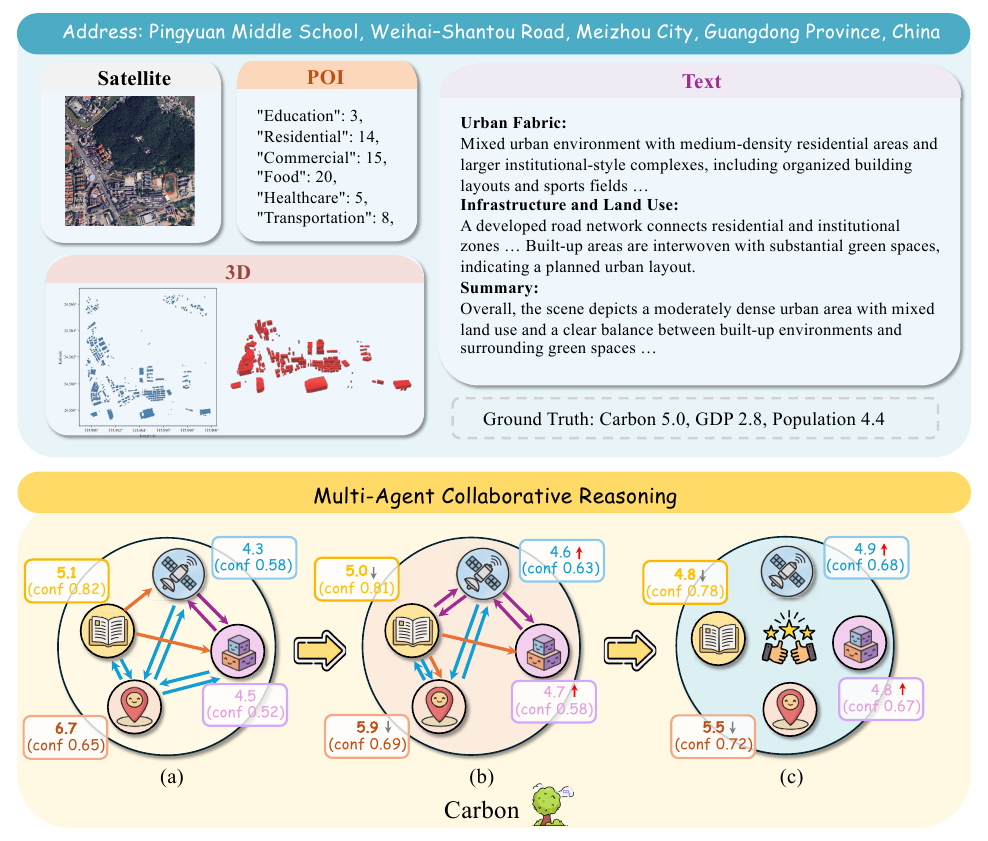}
  \vspace{-1.5em}
  \caption{Case study of Multi-Agent Collaborative Reasoning. Purple edges denote \textcolor{myPurple}{Corroboration}, blue edges indicate \textcolor{myblue}{Rectification}, and orange edges represent \textcolor{orange}{Supervisory} relations. Figure values represent the carbon indicator.}
  \label{fig:casestudy}
  \vspace{-1.5em}
\end{figure}

We present a representative example to illustrate Multi-Agent Collaborative Reasoning in Figure~\ref{fig:casestudy}.
As we can see, 
given a target region, the model first ingests multi-source information from satellite imagery, POIs, 3D building structures, and textual descriptions. Each modality-specific agent independently produces an initial prediction along with a confidence estimate. Due to the heterogeneous nature of the modalities, these initial predictions often exhibit notable discrepancies at the early stage (Figure~\ref{fig:casestudy} (a)).

Subsequently, the model proceeds through multiple rounds of multi-agent collaborative reasoning, where inter-agent interactions are explicitly modeled via a relation-aware collaboration mechanism. Consistent judgments are reinforced through corroboration (purple edges), conflicting judgments are questioned and corrected via rectification (blue edges), and agents with higher confidence exert guiding influence through supervision (orange edges).
During this iterative process, the predictions and confidence estimates of individual agents are progressively calibrated and driven toward convergence, with cross-modal inconsistencies transformed into signals that facilitate refinement. 
This process demonstrates \model’s ability to achieve more reliable urban indicator inference through explicit collaboration and systematic evidence integration.

\subsubsection{Visualization of Region Representations}
\begin{figure}[ht]
  \centering
  \includegraphics[width=\linewidth]
  {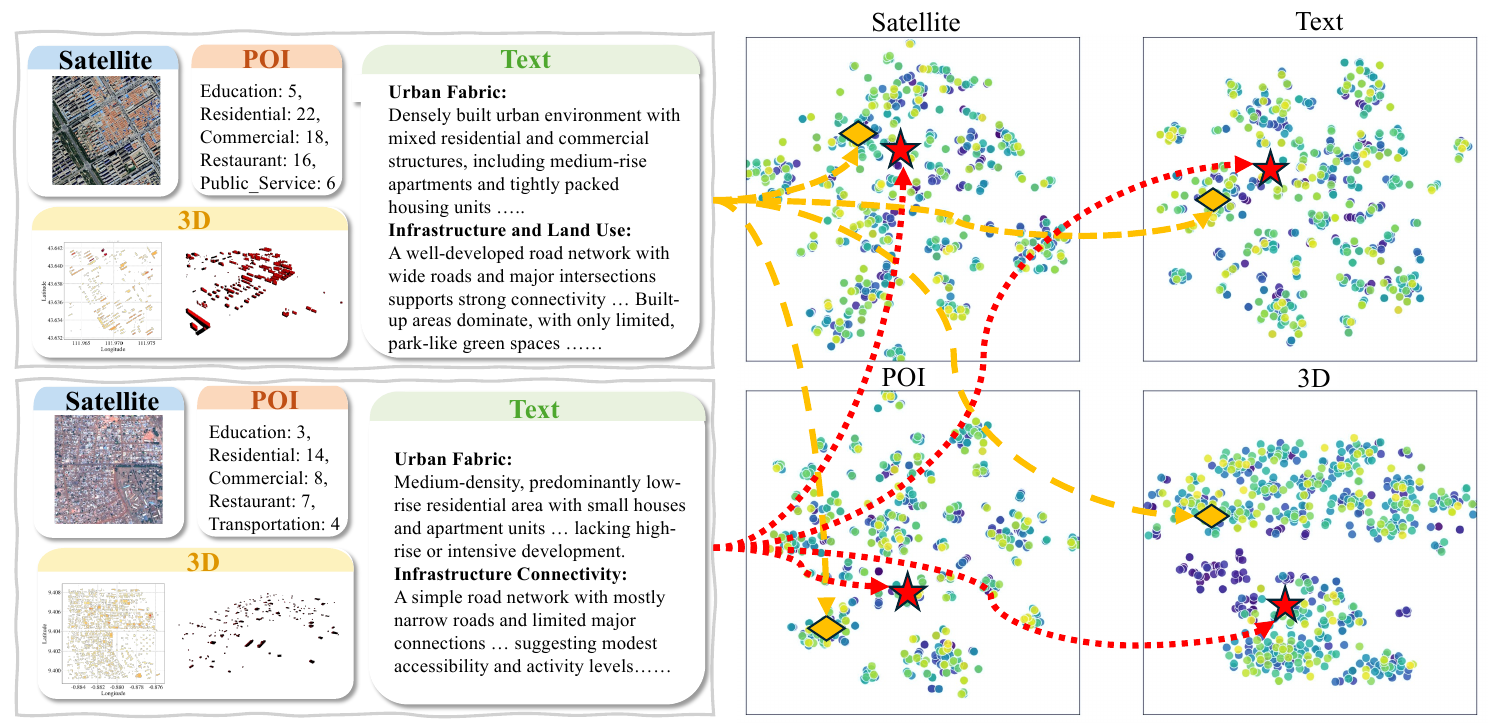}
  \caption{Representation space visualization.}
  \label{fig:vis_feat}
  \vspace{-1em}
\end{figure}

In this section, we visualize region representations using t-SNE in Figure ~\ref{fig:vis_feat}. The two regions exhibit markedly different distances across modality-specific feature spaces. From the satellite imagery perspective, they appear highly similar in urban morphology, characterized by dense built-up areas and regular road grids.
Although satellite-derived textual descriptions introduce additional external knowledge, the resulting representations remain predominantly driven by satellite imagery.

In contrast, the differences become more pronounced from the 3D perspective. One region exhibits a higher degree of building height variability and stronger mass concentration, indicating more intensive vertical development and land use. The other region, however, shows a flatter 3D distribution, with generally lower building heights and more dispersed volumes, lacking prominent vertical structural features and reflecting a development pattern dominated by low-rise or multi-story residential buildings. Such vertical morphological differences are often obscured in top-down satellite views but are explicitly amplified in the 3D representation space. This trend is also reflected in the relative distances observed in the 3D feature space, further highlighting the complementary role of 3D information in distinguishing urban regions with similar planar appearances.

\section{Conclusion and Future Work}

In this work, we propose \model, a tool-augmented multi-agent collaborative reasoning framework for urban region profiling, which advances beyond conventional DL-based profiling by enabling explicit multi-agent collaborative reasoning and evidence-grounded tool-use. By decomposing multimodal urban understanding into modality-specific agents and coordinating them through corroboration, rectification, and supervisory interactions, our model actively verifies, reconciles, and refines predictions to explicitly resolve modality inconsistency, rather than absorbing conflicts into a single latent fusion space. Moreover, RL-based agentic learning enables each modality agent to operate in a closed-loop “reason–act–observe” process: the agent can actively acquire external evidence and update its judgment iteratively, supported by curated tool-use trajectories and a tool-aware composite reward. 
Extensive experiments across multiple socioeconomic indicators demonstrate consistent performance gains over existing approaches.
Future work will extend \model~in three complementary directions: (i) incorporating richer external tools to provide more reliable and diverse evidence~\cite{feng2025earth}; (ii) scaling to a larger set of specialized agents to enable finer-grained division of labor; and (iii) exploring alternative interaction topologies to support more adaptive coordination and conflict resolution under heterogeneous evidence~\cite{feng2025earth,zhang2025gdesigner}.

\begin{acks}
This work is supported by the National Natural Science Foundation of China (No. 62402414), the Guangdong Basic and Applied Basic Research Foundation (No. 2025A1515011994), Guangdong Provincial Project 2025D03J0014, Guangzhou Municipal Science and Technology Project (No. 2023A03J0011), the Guangzhou Industrial Information and Intelligent Key Laboratory Project (No. 2024A03J0628), and Guangdong Provincial Key Lab of Integrated Communication, Sensing and Computation for Ubiquitous Internet of Things (No. 2023B1212010007).
\end{acks}
\clearpage

\bibliographystyle{ACM-Reference-Format}
\bibliography{reference}

\clearpage
\appendix

\twocolumn[{
    \renewcommand\twocolumn[1][]{#1}
    \begin{center}
      \textbf{\fontsize{15}{48}\selectfont Appendix}
    \end{center}
    \vspace{0.5cm}
}]

\section{Detailed Introduction of Tool-Use}
\label{appendix:tooluse}

In this section, we introduce the four kinds of external tools used in our Tool-Augmented Agentic Learning.
\begin{itemize}
    [leftmargin=*]
  \item \textbf{Web Search}. The model formulates a web search query to retrieve up to a fixed number of relevant socio-economic information sources from the internet. The search service is provided by a third-party web search service\footnote{\url{https://www.tavily.com}}. 
  The search request is formatted as: \texttt{\{name: search\_web, arguments: economic development, population statistics, industrial struc-\\ture energy consumption of "Address"\}}.
  The results consist of unstructured textual documents accompanied by their corresponding web URLs, which supply external and up-to-date contextual knowledge for evidence-based inference.

    \vspace{0.2em}
  \item \textbf{Crop-and-Zoom}.
    The model predicts a bounding box parameterized as $\texttt{bbox\_2d}$, specified by pixel coordinates on the original satellite image. This bounding box is used to crop and magnify a region of interest, enabling fine-grained visual inspection of local structures. The resulting observation is the zoomed-in cropped subfigure, which provides enhanced spatial details. This tool is only applicable to satellite imagery.
    \vspace{0.2em}
  \item \textbf{Nightlight Retrieval}.
    The model queries a nightlight intensity service using geographic coordinates (latitude and longitude) to obtain nighttime illumination measurements. 
    Higher intensities typically indicate greater levels of urbanization and economic activity. The observation is a scalar numerical signal that serves as a proxy indicator for regional development.
    \vspace{0.2em}
  \item \textbf{Historical Satellite Imagery Retrieval}.
    The model retrieves historical satellite imagery for the same geographic location at two time points (2015 and 2020) based on latitude and longitude. This tool enables temporal comparison of land-use patterns, urban expansion, and infrastructure development over an approximately five-year interval. 
    This tool is only applicable to satellite imagery.
\end{itemize}

\subsection{Discussion on Possible Web Search Data Leakage}

In our work, Web Search provides only qualitative contextual information (e.g., economic structure, industrial composition), not quantitative ground-truth labels. 
Our labels, including Carbon, Population, and GDP, are 1 km × 1 km gridded measurements derived from specialized remote sensing products, which are unavailable via general web search.
Web results typically appear as city-level aggregates with mismatched spatial granularity.

Furthermore, raw indicators are transformed into relative magnitude scores in [0, 9.9] based on the global dataset distribution, making direct label retrieval impossible even if a raw statistic were found online.

\section{Detailed Dataset Introduction}
\label{appendix:datasource}

\begin{figure}[ht]
  \centering
  \includegraphics[width=\linewidth]
  {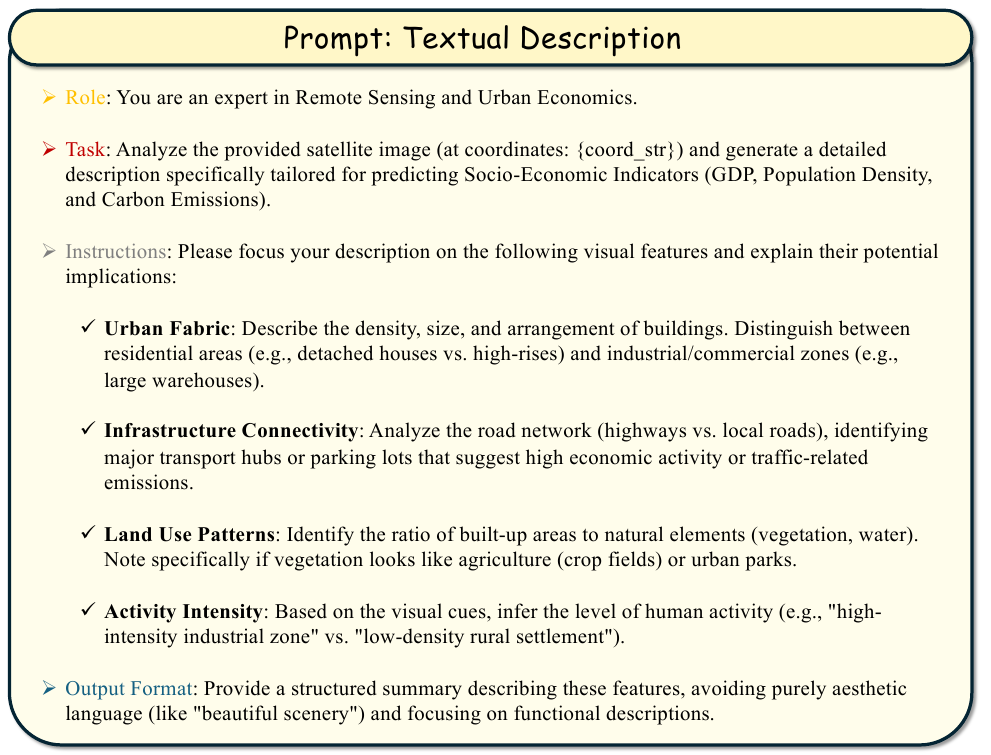}
  \vspace{-2.2em}
  \caption{Prompt for textual description.}
  \label{fig:text_prompt}
  \vspace{-1em}
\end{figure}

In this section, we introduction the data sources of our dataset including input modality data and downstream task data.
\begin{itemize}
    [leftmargin=*]
  \item \textbf{Satellite Imagery}. We collect satellite imagery from Google Maps using a tile-based retrieval scheme. For each region, we download an image centered at the region centroid (lat,lon) that covers an area of approximately 1km×1km. Specifically, we fix the zoom level to z=18, under which each map tile has a spatial resolution of 256×256 pixels, and stitch multiple tiles to obtain the target image. These images provide fine-grained visual cues related to urban morphology, land-use patterns, and built environment structures.
  \vspace{0.2em}
  \item \textbf{Textual Description}. Following~\cite{urbanclip, urbanvlp}, we generate region-level textual descriptions from satellite imagery using GPT-4o~\cite{hurst2024gpt4o}. These descriptions summarize salient visual patterns and inferred urban characteristics—such as building density, land-use functions, and infrastructure layouts—thereby serving as a complementary semantic modality to raw imagery. To ensure description quality and perceptual consistency, we retain only texts with a PerceptionScore~\cite{urbanvlp} above 0.6, filtering out low-confidence or visually misaligned generations.
  The prompt for generating textual description is provided in Figure~\ref{fig:text_prompt}.
  \vspace{0.2em}
  \item \textbf{POI}. We collect Points of Interest (POIs) data from OpenStreetMap (OSM) via the Overpass API\footnote{\url{http://overpass-api.de/api/interpreter}}
    . For each region, we query OSM entities within a circular area with radius=1000 meters centered at the region coordinates.
\textit{For incorporating POI data as input to the LLM, we construct a textual description based on category-level statistics, including the total number of elements and aggregated functional POI types (e.g., buildings, amenities, and land use)}.

    \vspace{0.2em}
  \item \textbf{3D}. We incorporate 3D building information from~\cite{3dglobfp}, which provides three-dimensional representations of urban built environments. This modality captures structural attributes such as building height distributions, volumetric density, and vertical spatial organization, complementing 2D satellite imagery with explicit geometric cues.
  We rasterize 3D building footprints with height attributes into a fixed 
64×64 height map using an affine transformation defined over the local bounding box~\cite{regiondcl}. From the rasterized map, we extract a structured textual representation that includes global height statistics (e.g., building coverage ratio, minimum/maximum/mean/median height), height distribution histograms (e.g., proportions of pixels falling into predefined height ranges), and a downsampled relative-height grid (e.g., a low-resolution normalized height map encoding spatial height patterns).
  \vspace{0.2em}
  \item \textbf{Carbon Emission}. Carbon emission data are sourced from the Open-source Data Inventory for Anthropogenic $\text{CO}_2$ (ODIAC) \footnote{\url{https://db.cger.nies.go.jp/dataset/ODIAC/DL_odiac2024.html}}
    . We use the 2023 release, which provides gridded fossil-fuel $\text{CO}_2$ emissions at a global scale. Monthly emission values are aggregated into an annual mean for each 1 km × 1 km region.
    \vspace{0.2em}
  \item \textbf{Population}. Following the setup in \cite{geollm}, Population data are obtained from the WorldPop project. We use the 2020 global unconstrained population mosaic, which provides gridded estimates of resident population at a spatial resolution of approximately 1 km × 1 km. Each grid cell stores the estimated population count within the corresponding area. Population values are aggregated to the region level and used as a key sociodemographic indicator.
  \item \textbf{GDP}. Regional economic activity is characterized using GDP data adapted from the CityLens dataset \cite{liu2025citylens} and publicly available records hosted on Zenodo\footnote{\url{https://zenodo.org/records/5880037}}
    . GDP values are reported in PPP-adjusted U.S. dollars (USD), representing the total gross domestic product aggregated within each 1 km × 1 km spatial grid cell. This modality provides a macro-level economic perspective complementary to fine-grained urban signals.

\end{itemize}

\section{In-depth Overview of Baseline Methods}
\label{appendix:baseline}

We provide a detailed introduction of baseline methods used in our paper.
\begin{itemize}
    [leftmargin=*]
  \item Qwen2.5-VL-3B~\cite{bai2025qwen25vl}: A lightweight 3B-parameter vision-language model in the Qwen2.5-VL family, designed to retain strong multimodal capability under resource constraints. It emphasizes omni-document parsing (e.g., handwriting, tables, charts), precise object grounding (via points/bounding boxes and structured coordinate / JSON formats), and ultra-long video understanding with second-level event localization, while also targeting agentic interaction on computers and mobile devices. 
Architecturally, Qwen2.5-VL follows a Vision Encoder + Vision-Language Merger + LLM design: the vision side is a redesigned ViT that supports native dynamic resolution, uses window attention for efficiency, and splits images into patches; for video, it introduces MRoPE aligned to absolute time to better model temporal dynamics and enable precise moment localization.
\vspace{0.2em}
  \item Qwen3-VL-4B~\cite{Bai2025Qwen3VL}: Qwen3-VL-4B is a lightweight yet strong vision–language foundation model in the Qwen3-VL family, built upon a 4B-parameter Qwen3 language backbone. It supports native interleaved multimodal inputs (text, images, and video) with long-context modeling up to 256K tokens, enabling robust cross-modal understanding and reasoning. Architecturally, it integrates a SigLIP-2 vision encoder with DeepStack multi-level feature fusion and interleaved MRoPE positional encoding, which jointly enhance fine-grained visual perception and vision–language alignment. Despite its compact scale, Qwen3-VL-4B demonstrates competitive performance on a wide range of multimodal benchmarks, making it a strong and efficient baseline for multimodal representation learning and reasoning tasks.
  \vspace{0.2em}
  \item InternVL2.5-4B~\cite{internvl25}: InternVL2.5-4B is a medium-scale multimodal large language model in the InternVL 2.5 series, following the widely adopted “ViT–MLP–LLM” architecture. It connects an InternViT-300M vision encoder with a 4B-level language backbone (e.g., Qwen2.5-3B-Instruct) via an MLP projector for efficient vision–language alignment. The model supports dynamic high-resolution visual inputs and is capable of processing single-image, multi-image, and video data. Through progressive training strategies and strict data quality filtering, InternVL2.5-4B achieves notably improved multimodal reasoning and understanding performance, offering a strong balance between efficiency and accuracy and serving as a competitive open-source baseline for multimodal representation learning.
  \vspace{0.2em}
  \item Kimi-VL-A3B~\cite{team2025kimivl}: Kimi-VL-A3B is an efficient open-source vision–language model built upon a Mixture-of-Experts (MoE) architecture, activating only 2.8B parameters in the language decoder while maintaining strong multimodal reasoning capabilities. It combines a native-resolution vision encoder (MoonViT) with an MoE language model, enabling effective understanding of high-resolution images, long videos, and long documents within a 128K context window. Kimi-VL-A3B demonstrates competitive performance across a wide range of vision–language benchmarks, including OCR, mathematical reasoning, multi-image understanding, and agent-based tasks, achieving a favorable balance between performance and computational efficiency and serving as a strong baseline for efficient multimodal representation learning.
  \vspace{0.2em}
  \item GPT-4o~\cite{hurst2024gpt4o}: GPT-4o is an end-to-end trained omni-modal large language model that natively supports text, image, audio, and video inputs within a unified architecture. Unlike prior modular multimodal systems, GPT-4o processes and generates different modalities using a single neural network, enabling stronger cross-modal alignment and reasoning. It achieves performance comparable to GPT-4 Turbo on English text and code, while significantly improving non-English, vision, and audio understanding, with substantially lower latency and cost. Owing to its strong multimodal perception and reasoning capabilities, GPT-4o serves as a powerful general-purpose multimodal foundation model and a representative upper-bound baseline for vision–language understanding and reasoning tasks.
  \vspace{0.2em}
  \item Gemini-2.5-Flash~\cite{comanici2025gemini25}: Gemini-2.5-Flash is a hybrid reasoning, natively multimodal large language model designed to balance strong reasoning capability with low latency and cost. It supports long-context inputs (up to 1M tokens) and multimodal understanding across text, images, audio, and video, while providing a controllable thinking budget that allows users to trade off reasoning depth against efficiency. Compared to flagship models, Gemini-2.5-Flash achieves competitive performance on reasoning, coding, and multimodal benchmarks at a fraction of the computational overhead, making it well suited as a scalable baseline for complex but cost-sensitive tasks.
  \vspace{0.2em}
    \item UrbanCLIP~\cite{urbanclip}: UrbanCLIP is the first framework that explicitly introduces text modality into urban region profiling by leveraging contrastive language–image pretraining. It generates high-quality region-level textual descriptions from satellite images using an MLLM, and jointly learns visual representations via image–text contrastive loss and language modeling loss. 
    \vspace{0.2em}
  \item UrbanVLP~\cite{urbanvlp}: UrbanVLP is a multi-granularity vision–language pretraining framework for urban socioeconomic indicator prediction. It jointly leverages macro-level satellite imagery and micro-level street-view imagery, and aligns them with high-quality textual descriptions through contrastive vision–language learning. Unlike UrbanCLIP, which relies on directly generated text from LLMs, UrbanVLP introduces an automatic text generation and calibration mechanism with a reference-free quality metric to mitigate hallucination and homogenization. By performing multi-granularity cross-modal alignment at both global and token levels, UrbanVLP learns more fine-grained and robust urban representations, achieving state-of-the-art performance and improved transferability across multiple urban tasks.
  \vspace{0.2em}
  \item AgentFusion: 
    AgentFusion aggregates the predictions of multiple modality-specific agents through MLP layers. 
    This approach does not model inter-agent communication or reasoning dynamics, but instead relies on supervised learning to capture fixed combination patterns across modalities, providing a straightforward parametric fusion baseline.
\vspace{0.2em}
\item AgentBlender: AgentBlender aggregates the outputs of four modality-specific agents using result-level fusion strategies, similar to LLM ensembling~\cite{llm-blender-2023}. We consider two variants: (i) \textit{confidence-weighted mean}, where agent predictions are combined via a weighted average based on their self-reported confidence scores; and (ii) \textit{majority voting}, where predictions are discretized into fixed-width bins and the final result is determined by cross-agent consensus. These variants serve as lightweight baselines for evaluating the effectiveness of collaborative reasoning mechanisms.
\item Agent-Arbiter: Agent-Arbiter is an LLM-based arbitration baseline that formulates multimodal fusion as a single-shot decision synthesis problem. For each sample, it inputs the predictions, self-reported confidence scores, and textual rationales independently produced by four modality-specific agents (satellite imagery, text, POI, and 3D structure), and employs an extra LLM as a judge to generate the final estimates for Carbon Emission, Population, and GDP. During arbitration, the judge explicitly considers cross-agent agreement and conflicts as well as confidence disparities, and outputs both the final predictions, calibrated confidence scores, and a natural-language justification for the decision.
\vspace{0.2em}
\item Hierarchical Prompting: Hierarchical Prompting organizes multiple modality-specific agents in a predefined sequential order, where agents operate one after another in a pipeline manner. The output of each agent (including predictions and intermediate reasoning) is passed as contextual input to the subsequent agent, allowing later agents to condition their predictions on earlier results. The agent order follows a progressive transition from directly observable urban form to higher-level semantic interpretation (Satellite $\rightarrow$ 3D $\rightarrow$ POI $\rightarrow$ Text), such that spatial and structural cues are established first, functional activity patterns are incorporated next, and textual reasoning is applied last for semantic synthesis and consistency checking. 

\end{itemize}

\section{Detailed Tool-Use Example Illustration}
\begin{figure*}[ht]
  \centering
  \includegraphics[width=\linewidth,
  ]
  {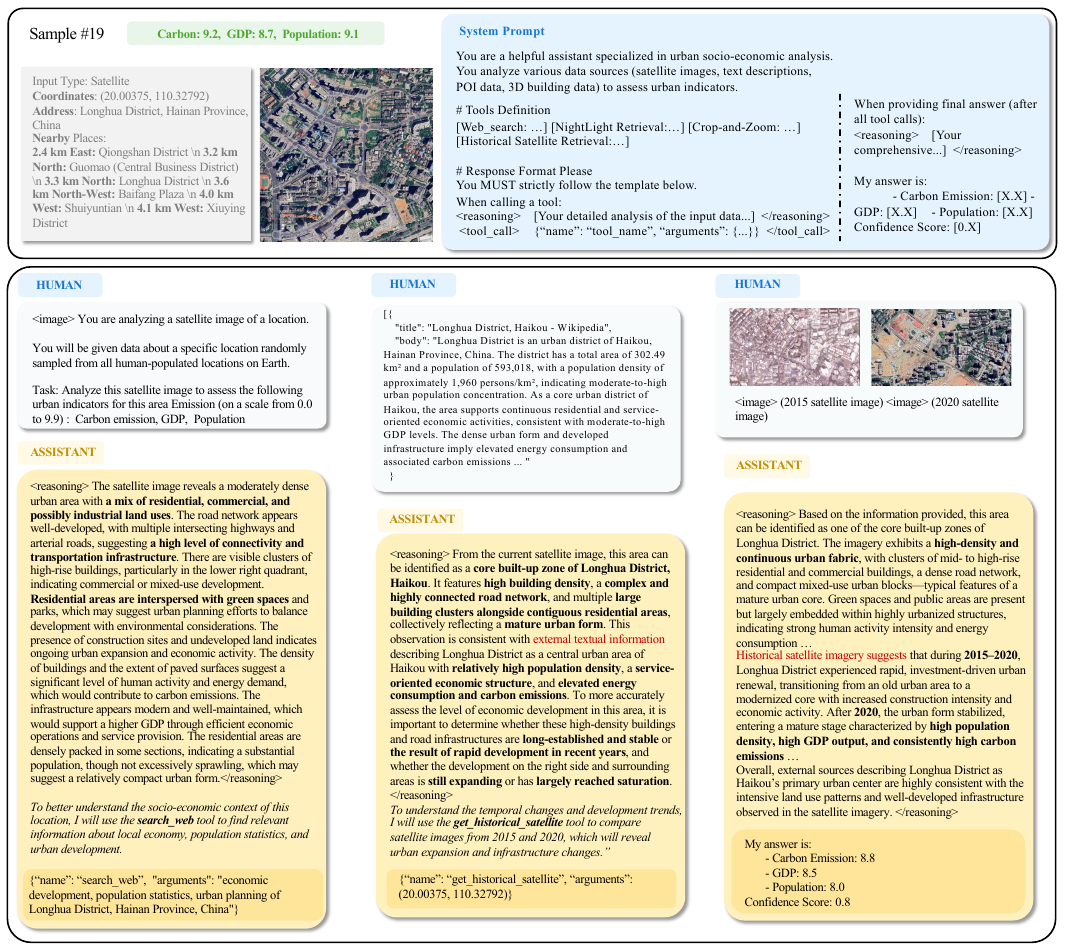}
  \vspace{-2em}
  \caption{Detailed tool-use example illustration.}
  \label{fig:detail_tooluse_example}
\end{figure*}

Figure~\ref{fig:detail_tooluse_example} demonstrates detailed Tool-Use Example corresponding Figure~\ref{fig:tooluse} in the main text.

\section{Discussion about Evaluation Metrics}
We justify our choice of $R^2$ and Spearman $\rho$ over RMSE from three perspectives: 
\textbf{(1) Limited interpretability of RMSE after label scaling.}
Following~\cite{geollm,urbanr1,liu2025cityrise}, we convert raw indicator values into rank-normalized scores in $[0, 9.9]$. RMSE computed on scaled values loses clear physical meaning, and since this transformation is non-invertible, scaling back to original values for recomputation is infeasible. By contrast, $R^2$ remains interpretable as the proportion of variance explained, regardless of the scaling procedure.
\textbf{(2) Suitability for rank-normalized labels.}
Since the scaled labels reflect relative ordering rather than absolute
magnitude, Spearman $\rho$ naturally measures rank consistency without
depending on absolute values. Together, $R^2$ and Spearman $\rho$
provide complementary evaluation: the former captures variance
explanation, while the latter assesses ranking consistency.
\textbf{(3) Consistency with prior work.}
Related studies adopting similar label-scaling protocols~\cite{geollm,urbanr1,liu2025cityrise} do not report RMSE, and we follow the same evaluation protocol to ensure fair and comparable assessment.

\end{document}